# Towards Self-Evolving Agents: A Human-Inspired Adaptive Exploration-Exploitation Framework for Genetic Network Programming

**Ali Kohan [1], Mohamad Roshanzamir [1,*], Roohallah Alizadehsani [2,3], Seyedali Mirjalili [4,5]**

[1] Department of Computer Engineering, Faculty of Engineering, Fasa University, Fasa 74617-81189, Iran

a-kohan971@fasau.ac.ir

roshanzamir@fasau.ac.ir

[2] Institute for Intelligent Systems Research and Innovation (IISRI), Deakin University, Geelong, Australia

[3] Center of Excellence in Precision Medicine and Digital Health, Department of Physiology, Faculty of Dentistry, Chulalongkorn University, Bangkok 10330, Thailand.

r.alizadehsani@deakin.edu.au

[4] Centre for Artificial Intelligence Research and Optimization, Torrens University Australia, Brisbane, QLD 4006, Australia

[5] University Research and Innovation Center, Obuda University, Budapest, Hungary

ali.mirjalili@torrens.edu.au

**corresponding author: Mohamad Roshanzamir*

# Abstract

Recent advancements in agentic AI have increasingly moved toward graph-based methods, driven by the demand for explainable, human-centered, and non-linear reasoning workflows. A prominent example is Genetic Network Programming (GNP), a self-evolving algorithm that utilizes directed graphs to evolve interpretable decision structures for agents. As in most evolutionary algorithms, effectively balancing exploration and exploitation is a key aspect of GNP. However, this trade-off has received limited attention in the GNP literature. To address this gap, we draw inspiration from human developmental patterns, where children prioritize broad experimentation and *action* over *deliberation*, with this tendency reversing with age. By mapping transitions between GNP's judgment nodes to *deliberation* and processing nodes to *action*, we propose Human-Inspired GNP (HGNP), a novel adaptive framework that dynamically regulates the exploration–exploitation balance throughout the evolutionary process. The method consists of novel adaptive crossover and mutation operators, and a cycle elimination mechanism. HGNP not only improves the evolutionary process but also provides a framework for adjusting the exploration-exploitation balance based on the characteristics of the target environment and its search space. This approach is more effective than tuning via crossover and mutation probabilities in standard GNP. The modifications are general and can be applied to almost all GNP variants. When integrated with standard GNP and two recently introduced GNP variants and evaluated on the Tileworld benchmark, HGNP demonstrated significant performance improvement in agents' strategy. The combination of HGNP with Situation-based GNP (HGNP-SBGNP) achieved the best overall results.

# 1. Introduction

The rapid evolution of agentic Artificial Intelligence (AI) has redefined the landscape of autonomous decision-making, shifting the focus from passive models to active agents capable of goal-directed planning and complex decision-making [1]. A notable development in this progress is self-evolving agents, which can iteratively refine their internal logic [2, 3]. As these systems are increasingly integrated into critical sectors, the demand for Explainable AI (XAI) [4-6] and trustworthy agentic workflows has become paramount [7]. While Deep Learning (DL) and Large Language Models (LLMs) have significantly advanced agentic capabilities, they often function as "black boxes" and struggle with reliable planning and structural transparency. To address these limitations, graph-based representations have emerged as compelling alternatives, offering the interpretable workflows and non-linear reasoning required for robust agentic behavior.

Another fundamental challenge in the development of self-evolving agentic systems is managing the exploration–exploitation trade-off [8-10]. To remain effective, an agent must balance exploring new strategies to handle environmental shifts with exploiting known, high-quality behaviors to ensure performance. This tension is closely linked to the conflict between interpretability and performance: while sophisticated methodologies such as Reinforcement Learning (RL) offer powerful techniques for dynamically adapting this balance, they often lack structural transparency and struggle with high computational overhead [11]. In contrast, graph-based evolutionary frameworks such as Genetic Network Programming (GNP) [12] provide a unique middle ground between interpretable symbolic agents and adaptive learning systems and are suitable for low-resource environments. However, GNP's literature has largely overlooked the dynamic adjustment of the exploration–exploitation trade-off. Most existing GNP variants rely on fixed parameters or simple heuristic adjustments, failing to capture the dynamic complexity required for advanced agentic autonomy.

GNP is an Evolutionary Algorithm (EA) that extends conventional EAs, i.e., the bit-string structure of standard Genetic Algorithm (GA) and tree structure of standard Genetic Programming (GP), to a directed graph structure. This algorithm develops a distinguished directed graph structure to model its representation, making it practical for self-evolving agentic systems. It generates symbolic models, often more interpretable than complex black-box models like deep neural networks [13]. These symbolic models can be analyzed and understood, making them suitable for high-stake applications. For example, a GNP-generated graph can be broken down into clear, human-readable steps. The graph structure of GNP allows it to handle complex relationships, enabling efficient problem-solving for decision-making processes and interactions between multiple agents. Using a graph structure means that nodes in GNP can be reused multiple times during task execution. This reusability leads to compact and efficient solutions, as the same nodes can be used several times without replication. GNP is highly adaptable to various problem domains. Its directed graph structure, consisting of judgment and processing nodes, enables the flexible representation of "IF-THEN" decision rules, making this algorithm suitable for representing complex decision-making processes. It is suitable for use in a wide range of agent control problems [14-16], data mining problems [17, 18], and even intrusion detection problems [19-21].

GNP holds significant potential for Industry 5.0 by enhancing human-machine collaboration, enabling human-centered agentic processes within complex industrial environments. Unlike Industry 4.0's emphasis on pure automation, Industry 5.0 focuses on the synergy between human intelligence and AI [22-25]. In this context, AI systems are designed not only to perform tasks autonomously but also to work alongside humans in a way that enhances human capabilities, allowing for more meaningful human input and oversight. A central aspect of Industry 5.0 is its focus on explainable and human-centered AI, where decision-making processes are transparent, and human operators can understand, modify, or intervene in AI-driven actions [26]. GNP aligns well with these goals, as its graph-based structure is not only interpretable but also inherently modifiable by human users. This dual characteristic of GNP, i.e., being both understandable and modifiable, enables human operators to observe the decision-making processes, adjust the network's structure, and intervene in real time.

This paper introduces Human-Inspired GNP (HGNP), a novel method inspired by human decision-making that enhances GNP performance by dynamically managing the exploration-exploitation trade-off. In summary, our main contributions are:

- Human-Inspired Adaptation: Drawing inspiration from human decision-making patterns to enhance the exploration-exploitation balance in GNP.
- Novel Operators: Introducing adaptive crossover and mutation operators along with a cycle elimination method to improve GNP's evolutionary process.
- Dynamic Trade-off Adjustment: Developing a framework that allows fine-tuning exploration and exploitation beyond traditional probability-based adjustments.
- Generalizability: Designing the proposed modifications to apply broadly across different GNP variants.
- Performance Improvement: Demonstrating significant enhancements in GNP and two of its recent variants on the Tileworld benchmark.

The rest of this paper is organized as follows. Section 2 provides background by reviewing prominent versions and applications of GNP and graph-based agentic AI. Section 3 briefly introduces GNP and its evolutionary mechanisms, i.e., crossover and mutations. Section 4 describes the proposed mechanism of HGNP in detail. Section 5 evaluates this work in a benchmark testbed, and includes a discussion of the results. Finally, in Section 6, the conclusions and future work are presented.

# 2. Related Works

The literature surrounding interpretable agents and graph-structured decision-making can be broadly categorized into foundational algorithmic developments and emerging paradigms in agentic AI. This section provides a comprehensive review of these trends, organized into three primary areas. First, we examine the core advancements in GNP (Section 2.1), detailing its practical applications, RL-based integrations, probabilistic modeling, and the evolution of its genetic operators. Second, we explore the recent industry-wide shift toward graph-based reasoning in LLM agents (Section 2.2), which provides a contemporary context for the structural advantages of graph-based decision frameworks. Finally, we synthesize the limitations identified in both traditional and modern approaches to establish the fundamental research motivation and the specific gaps addressed by this work (Section 2.3).

## 2.1. Genetic Network Programming (GNP)

### 2.1.1. Applications

GNP has been frequently used for data mining and association rule discovery. The method proposed in [27] aimed to generate association rules from large, densely populated datasets using GNP while accounting for a real-world database with a significant number of attributes. A large database was divided into multiple smaller datasets, with each GNP handling one dataset containing a suitable number of attributes. These GNPs were then processed in parallel. Additionally, several new genetic operations were introduced to improve the extracted rules' quantity and quality. Madokoro et al. [28] proposed a filter creation method by modifying sensor signals using GNP. This method aims to enable automatic calibration to account for individual variations. This approach primarily focuses on acquiring knowledge about the topology and the relationships between input features and the signals used for teaching. In [18], two types of classifiers were introduced using class association rules. One classifier relies on semisupervised learning, while the other combines supervised and semisupervised learning. The latter method creates multiple classifiers using these learning techniques and reaches a final decision by merging the classification outcomes from these classifiers. The study employs the GNP method for association rule mining, aiming to broaden the use of GNP in data mining that incorporates semisupervised learning. In [29], a comprehensive approach was applied, incorporating the GNP, RL, and a multi-layer perceptron to classify data and forecast stock returns. The GNP's outcome was used for return prediction through accumulation rules. The purpose of integrating these outputs was to estimate the one-day return. Köhnke and Borgelt [30] introduced a variable-size GNP extension featuring a novel mutation operator that allows individuals to dynamically adjust the number of nodes to match task complexity without causing the bloat problem. The proposed approach improves flexibility and performance in high-dimensional tasks, demonstrated through enhanced results in financial portfolio optimization while preserving interpretability.

### 2.1.2. RL-based

In some studies, RL [31] enables GNP to adapt and learn more efficiently from its interactions with the environment, thereby enhancing its decision-making abilities. This version of GNP is used in some research, such as [32-36]. In [32-35], the learning phase of GNP employs the SARSA algorithm [31]. The algorithm utilized in [36] supports online learning, enabling programs to be incrementally modified based on rewards gained during task execution. This approach is applied to stock trading rules, mainly through technical analysis, to determine optimal buy and sell times based on technical indicators. Niching Genetic Network Programming (GNP-Niching) [37] allows GNP to preserve diversity within the solutions. It avoids premature convergence and facilitates discovering multiple optimal solutions in a multi-modal problem space. A new learning classifier system (LCS) called *niching GNP with rule accumulation* was used in [38]. This method stands out due to three main features. First, it uses GNP as the rule generator, which enhances knowledge representation more effectively than traditional GA-based LCSs. Second, GNP includes an innovative niching mechanism that encourages the discovery of diverse and high-quality rules. Finally, it incorporates an RL mechanism to assign credit precisely to the discovered rules.

### 2.1.3. Probabilistic Modeling

Diversity preservation and probabilistic model building in GNP aim to enhance evolutionary performance by maintaining population diversity and guiding the search process. In [39], a new evolutionary approach was proposed and applied to address traffic prediction challenges using class association rule mining. This study develops a probabilistic model by estimating the probability distribution based on selected elite individuals from the previous generation. This probabilistic model enhances the evolutionary process and helps achieve the desired goal. The paper introduced two methods focused on extracting probabilistic information about node connections and node transitions to construct the probabilistic model. Li et al. [40] proposed a novel algorithm called probabilistic model building genetic network programming (PMBGNP), which leverages GNP's graph structure to offer improved solution representation over traditional estimation of distribution algorithms. To further refine this, a reinforced version of PMBGNP was developed, combining reinforcement learning with the original framework. This integration improves the system's fitness outcomes, search efficiency, and reliability when controlling agents' behavior. A similar study presented in [16] introduces a new method that utilizes undesirable individuals by focusing on their substructures instead of their entire structures to address RL problems. The effectiveness of this approach is demonstrated through its application to an RL problem, particularly in robot control problems. Li et al. [41] extended PMBGNP from a discrete to a continuous search space. Beyond applying the traditional PMBGNP method to optimize node connections in graph structures, they introduced a Gaussian distribution to model the continuous variables of the nodes. This approach mirrors the classical continuous population-based incremental learning, where the mean and standard deviation are calculated accordingly. The actor-critic RL technique updated these mean and standard deviation values. This algorithm was then applied to a particular RL problem related to autonomous robot control, where the robot's wheel speeds and sensor readings are continuous variables.

#### 2.1.4. Enhancing Operators

In [42], an adaptive GNP is introduced, which dynamically adjusts the crossover and mutation probabilities based on the current conditions. Adaptive GNP enables the algorithm to modify its parameters in real-time, improving its capacity to adapt to changing environments or problem requirements. This adaptive feature allows the system to shift the evolutionary focus towards the nodes frequently reused in high-quality solutions. However, this frequency-based adaptation may overlook less frequently used nodes that are still critical to the solution, leading to their potential disruption. Additionally, its adaptivity is limited to conventional genetic operators, which restricts its potential to efficiently balance exploration and exploitation. The researchers in [43] presented a novel approach to GNP by integrating an ant colony mechanism to enhance performance in deterministic and stochastic environments. The proposed method employs a constructive process that preserves beneficial structures across generations, improving decision-making strategies. The results demonstrate that this new approach significantly enhances the algorithm's efficiency, particularly in uncertain environments. This study highlights the potential of combining ant colony optimization with GNP to solve complex agent control problems effectively. A similar work is done in [44] in which the researchers introduced a two-phase approach for exploration and exploitation, allowing for a balance between generating diverse solutions and refining promising individuals. By leveraging the accumulated experience of successful individuals, the algorithm aims to enhance the efficiency of GNP. This method has a significant drawback: calculating accumulated experience in successful individuals is time-consuming, causing the algorithm to be slow.

A new version of GNP, called situation-based GNP (SB-GNP), was introduced in [45]. SB-GNP improves upon traditional GNP by allowing each agent to develop its strategy based on its specific situation, thus optimizing resource use and reducing complexity. This method addresses the challenges of resource limitations in multi-agent systems, enabling faster and more effective learning. In a recent study, Kohan et al. [46] aimed to reduce the search space of SB-GNP by employing simplified operators. Since SB-GNP has a larger search space compared to standard GNP, the use of these simplified operators leads to improved convergence in such extensive search spaces. Parallel GNP [47] involves functionally distributed systems comprising multiple tasks. It utilizes parallel computing techniques to enhance the efficiency and scalability of GNP, making it well-suited for tackling large-scale and complex challenges. In Parallel GNP, the GNPs associated with these tasks operate autonomously and concurrently, effectively resolving conflicts in task execution. Compared to traditional GNP, Parallel GNP exhibits faster convergence and better fitness results, as demonstrated by simulations performed on dynamic problems.

### 2.2. Graph-Based Reasoning in LLM Agents

Recent advancements in agentic AI have increasingly moved toward graph-structured methods to overcome the linear reasoning limitations of LLMs and LLM agents. While most LLMs are constrained in several core agentic procedures (e.g., such as reliable planning, long-term memory, tool management, and multi-agent coordination), graphs can function as a powerful structure to improve organization and coordination in complex agentic workflows. This has led to a growing trend toward Graph-augmented LLM Agents (GLA) to help bridge the gap between high-level reasoning and executable task planning [48, 49].

For instance, AgentKit [50] provides a modular architecture where sub-tasks are represented as nodes, enabling the construction of explicit "thought processes" through dynamic directed graphs. Similarly, Graph-of-Thought (Graph-CoT) frameworks like GLM [51] and Graph Counselor [52] utilize multi-agent synergies—partitioning tasks among specialized agents for planning, retrieval, and execution—to navigate knowledge graphs and improve the factual accuracy of complex reasoning. GraphAgent [53] further extends this by integrating graph generators and task-planning agents to handle both explicit and implicit semantic dependencies. These studies underscore a growing industry consensus that interpretable, graph-based structures are essential for managing the multi-step dependencies required for agentic workflows.

Furthermore, the challenge of efficiently navigating these graph-based decision spaces has led to the integration of graph learning and distributed computation theory. Wu et al. [54] demonstrated that while standard LLMs often struggle with graph navigation due to auto-regressive biases, integrating Graph Neural Networks (GNNs) can significantly improve task planning. To address scalability, frameworks like GraphAgent-Reasoner [55] apply distributed graph computation to decompose global reasoning problems into node-centric tasks, while AgentRouter [56] utilizes knowledge-graph-guided routing to select the most appropriate agent for a given query. The integration of evolutionary search into modern agentic frameworks is further exemplified by EvoFlow [57], which employs a niching evolutionary algorithm to automatically evolve populations of diverse and complexity-adaptive agentic workflows. By applying crossover and mutation operators to optimize graph-structured LLM interactions, this approach demonstrates that evolutionary principles remain a powerful and relevant tool for managing the structural complexity of contemporary multi-agent planning systems.

These developments highlight a fundamental shift toward graph representations for managing complex reasoning and planning in modern agentic systems. By structuring tasks, intermediate states, and dependencies as interconnected nodes and edges, these approaches enable agents to move beyond linear, step-by-step reasoning and instead operate over richer, more flexible computational structures. Graph-based structures not only improve the organization of complex decision workflows but also enhance the interpretability of the process [50, 58]. Despite these advantages, increasing structural complexity makes efficient convergence a major bottleneck, particularly in large-scale and highly dynamic environments.

### 2.3. Research Motivation

GNP is introduced as a promising evolutionary computing technique that enhances the transparency and interpretability of agentic AI systems [13]. However, in most of these studies, GNP typically employs standard crossover and mutation techniques to evolve network structures, such as the widely used uniform genetic operators [59] or more advanced simplified genetic operators [60, 61]. While in traditional evolutionary algorithms, crossover and mutation operators are used for exploitation and exploration, respectively, they are not specifically optimized for GNP-related challenges to adjust this trade-off effectively. Current adaptive GNP methodologies often rely on simple probabilistic adjustments of standard genetic operators, which limits their effectiveness in complex environments. To address this, a more robust framework is required that utilizes a crossover operator optimized for exploitation and a more purposeful exploration beyond blindly stochastic gene alterations. This paper presents methods to enhance both exploration and exploitation in GNP, offering an adaptive framework to balance this trade-off optimized for GNP's graph structure and its specific challenges.

## 3. Overview of GNP

GNP is an evolutionary computation technique that extends GA and GP. While GAs and GPs primarily work with linear structures (strings or trees), the GNP algorithm operates with graph-based structures, enabling it to better model complex systems and decision-making processes. Figure 1 displays an example of the phenotype and genotype of an individual in GNP.

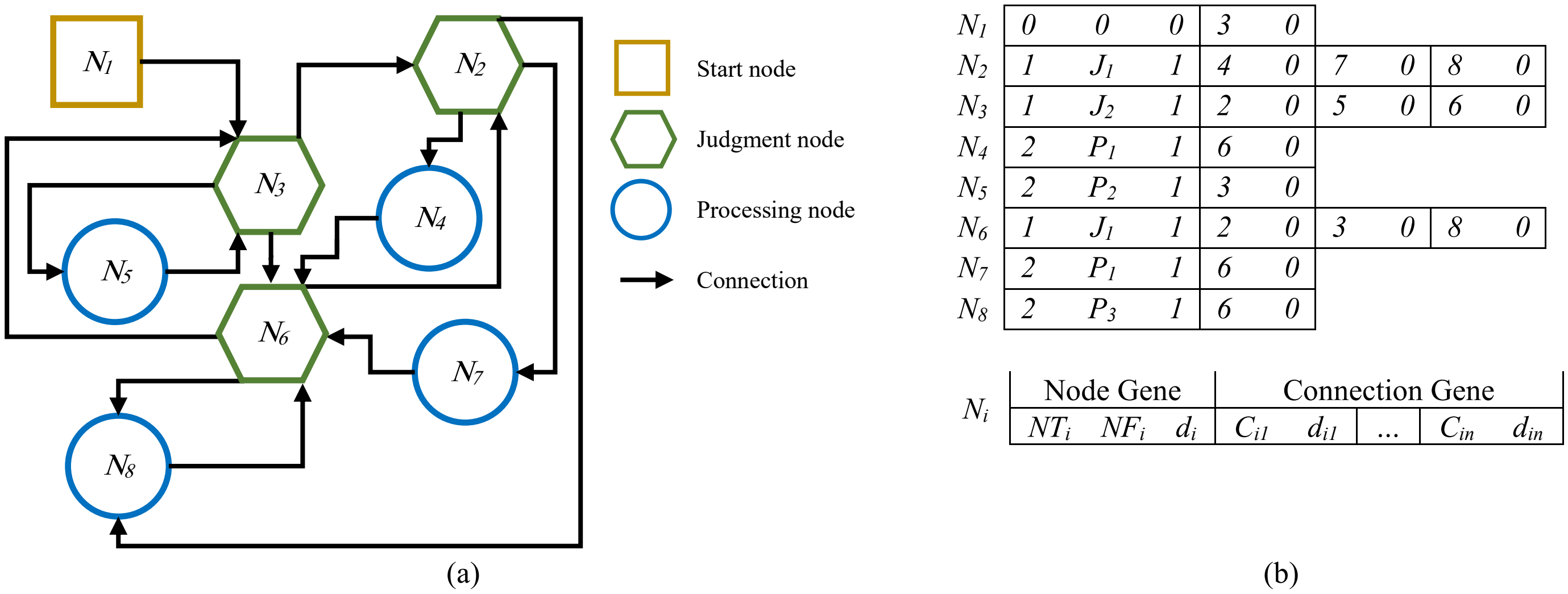

| $N_1$ | 0 | 0 | 0 | 3 | 0 | | | | |
|---|---|---|---|---|---|---|---|---|---|
| $N_2$ | 1 | $J_1$ | 1 | 4 | 0 | 7 | 0 | 8 | 0 |
| $N_3$ | 1 | $J_2$ | 1 | 2 | 0 | 5 | 0 | 6 | 0 |
| $N_4$ | 2 | $P_1$ | 1 | 6 | 0 | | | | |
| $N_5$ | 2 | $P_2$ | 1 | 3 | 0 | | | | |
| $N_6$ | 1 | $J_1$ | 1 | 2 | 0 | 3 | 0 | 8 | 0 |
| $N_7$ | 2 | $P_1$ | 1 | 6 | 0 | | | | |
| $N_8$ | 2 | $P_3$ | 1 | 6 | 0 | | | | |

*Figure 1. An example of an individual in GNP in a) phenotype and b) genotype forms.*

In this figure, $N_i$ shows the node ID. $NT_i$ shows the node type. Indeed, 0, 1, and 2 show the start, judgment, and processing nodes, respectively. $NF_i$ indicates which function should be executed, and $d_i$ indicates how long it takes the $NF_i$ function to be executed. Finally, $c_{ij}$ indicates which nodes $N_i$ is connected to. If the node type is judgment, there will be more than one output connection according to each condition.

As shown in Figure 2 (a), for crossover in the standard GNP algorithm, two parents exchange the connections of the corresponding nodes with probability $P_c$. This probability is equal for all nodes and in all iterations. Mutation is applied to an individual by selecting connections with constant probability $P_m$ and randomly changing them to generate offspring, as illustrated in Figure 2 (b). Like $P_c$, $P_m$ is equal for all nodes in all iterations.

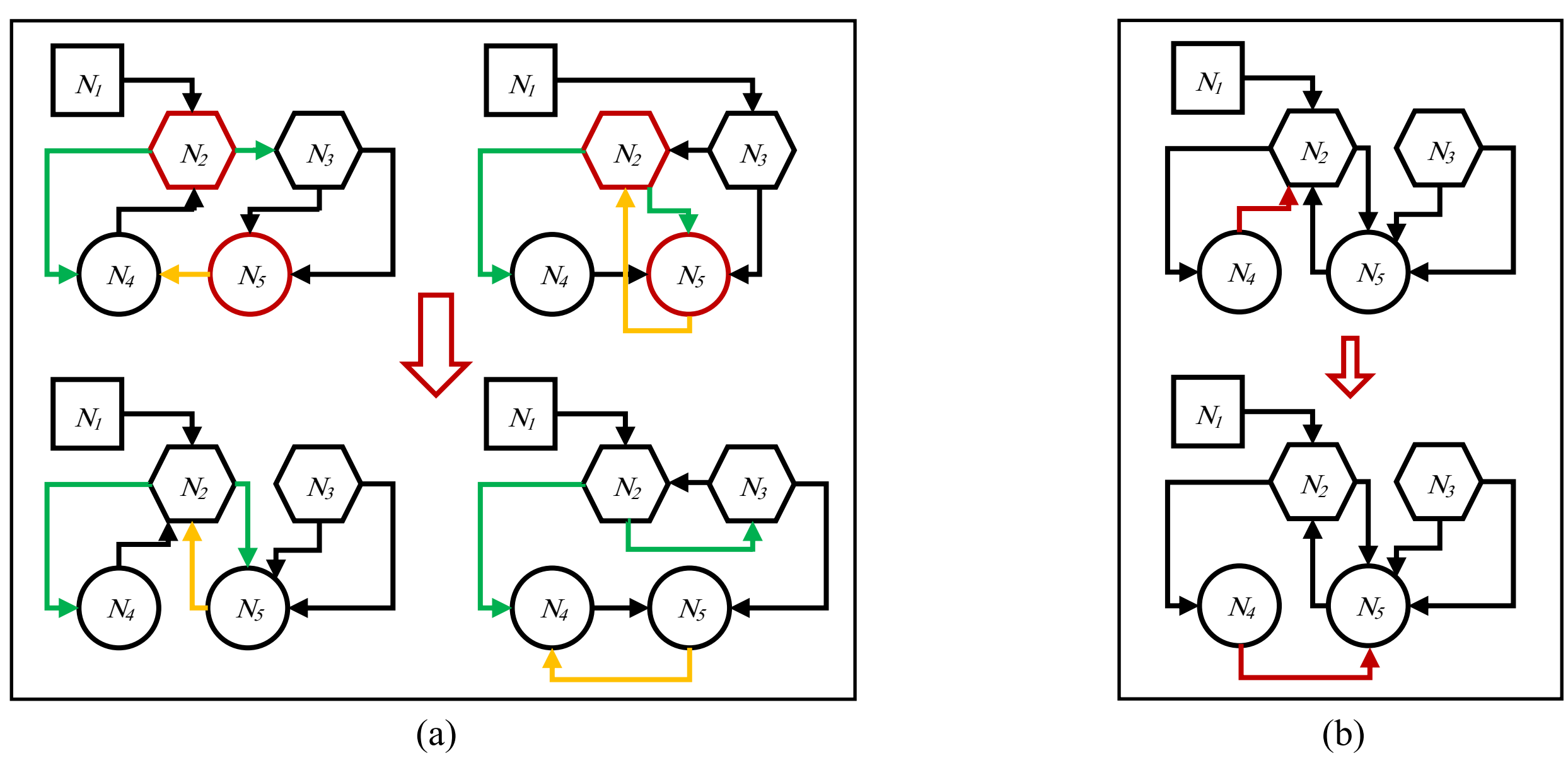

*Figure 2. An example of a) crossover and b) mutation operators in GNP.*

# 4. Proposed Methods

In evolutionary algorithms, it is generally better to start the algorithms by exploring the environment. Then, based on previous experiments, the algorithms tend to employ more exploitation strategies. However, as discussed in section 2, standard GNP and most of its variants lack a dynamic mechanism to adjust this balance, maintaining a relatively static relationship between exploration and exploitation throughout the evolutionary process. We introduce Human-Inspired GNP (HGNP) to change this mechanism, incorporating three new modifications to improve the evolution process. The first modification focuses on modifying the exploitation mechanism, while the others enhance the exploration mechanisms.

These modifications propose a new framework leading to improved performance compared to standard GNP, where the exploration-exploitation trade-off is controlled solely by adjusting $P_c$ and $P_m$. The overall workflow of these modifications is illustrated in Figure 3. The traditional mutation stage is retained to preserve the unrestricted exploration capability provided by standard mutation. They are explained in detail in the following sections.

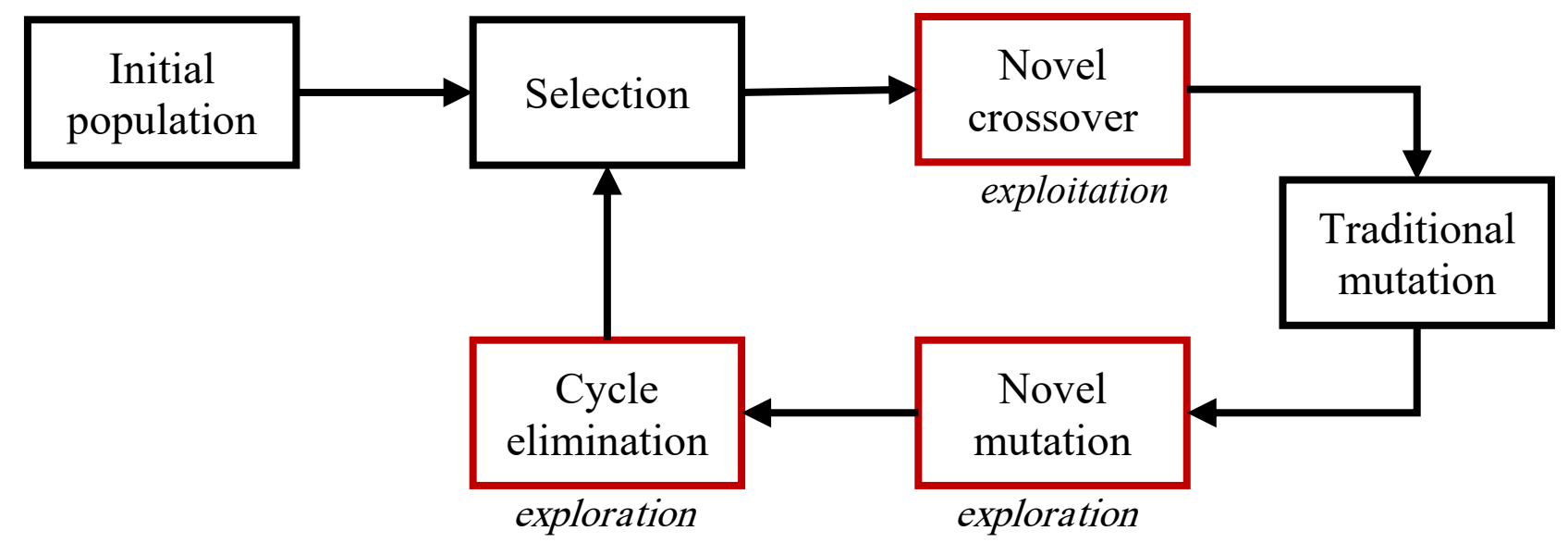

*Figure 3. Overview of the proposed methods. The modified operators are highlighted in red.*

## 4.1. Exploitation Phase

As mentioned previously, in GNP, the agents operate according to the structure of a graph. In this graph, nodes with more input connections are expected to influence the agents' actions more because they serve as central hubs. In a directed graph representation, a high in-degree indicates that a significant number of potential decision sequences transition through that specific node. We refer to these nodes as influential nodes. In the early iterations, there is no difference between nodes because of the random generation of the individuals. However, as the graph evolves, the difference between nodes becomes more pronounced. Changing the connections of influential nodes leads to a substantial change in the agents' behavior and often causes a decrease in exploitation and a delay in convergence. To address this problem, we need to decrease the probability of changing the influential nodes' connections as we approach the endpoint.

### 4.1.1. Crossover

In standard GNP, all nodes have an equal probability of being selected in the crossover operation. This probability is often denoted as $P_c$. To decrease the probability of changing the output connections of influential nodes, we need to reduce this probability proportional to the number of input (entrance) connections of each node. Figure 4 displays an example of counting the number of input connections for each node in phenotype form. Nodes with thicker borders indicate higher input connection counts. In genotype form, the number of input connections is calculated by counting how often the node index appears in the connection genes. In the proposed method, an adaptive crossover probability is computed based on the number of input connections (i.e., the influence of the node) and the iteration number. This approach reduces the selection probability of highly influential nodes, thereby increasing exploitation by preserving their output connections. To calculate the adaptive crossover probability, two individuals are selected as parents. In these parents, the connections of the corresponding nodes $n$ are swapped with probability $P_c(n)$. To calculate $P_c(n)$, as the first step, the number of input connections for each node is normalized into [0, 1] using Equation 1:

$$W_i = \frac{IC_i - IC_{min}}{IC_{max} - IC_{min}} \quad \forall i \in \{nodes\ of\ the\ individual\} \tag{1}$$

where $W_i$ is the importance value of node $I$ normalized into [0, 1]. $IC_i$ is the number of input connections for node $i$. $IC_{min}$ and $IC_{max}$ are the minimum and maximum number of input connections across all nodes, respectively.

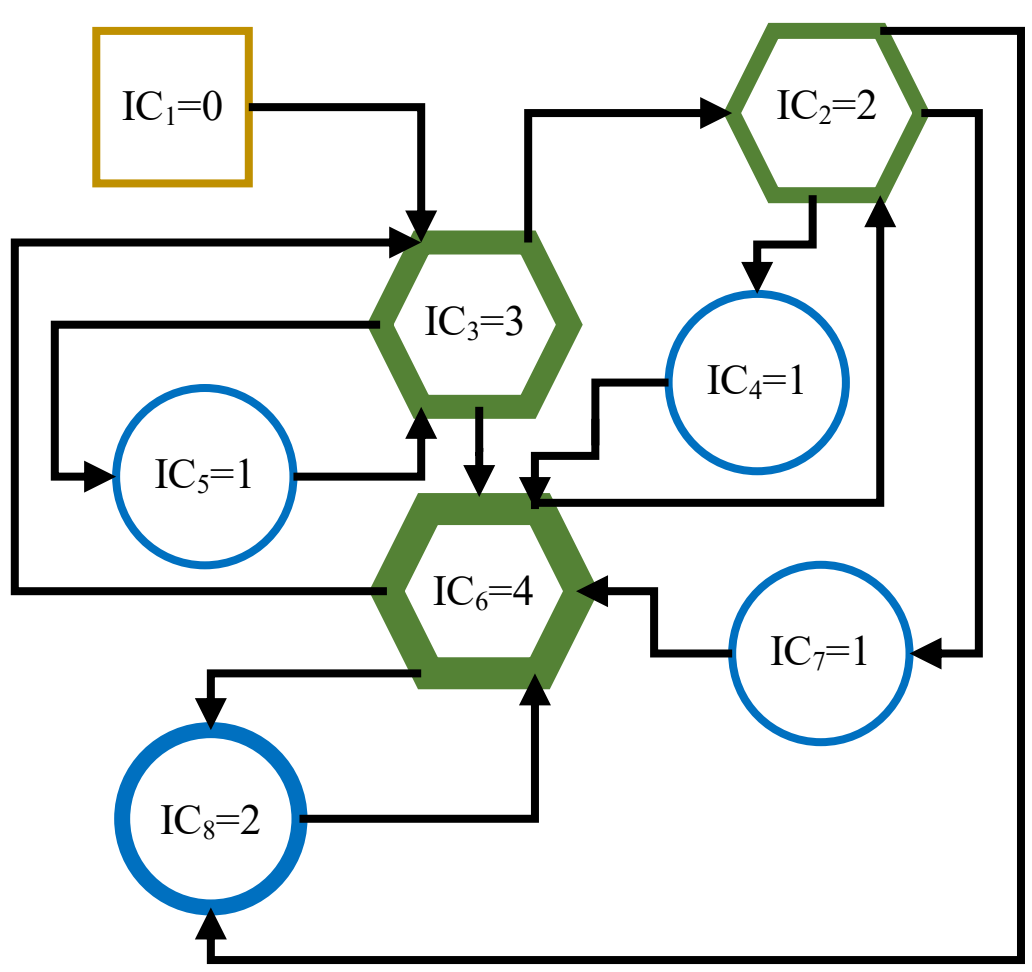

*Figure 4. An example of the number of input connections in phenotype form.*

Normalization is used to scale the parameter to a consistent range [0, 1], providing balanced influence and stability in the equation's computation. This process is essential in ensuring that all parameters contribute proportionally, regardless of their original scale, thereby preventing any single parameter from dominating the computation. Additionally, normalization reduces the sensitivity of the equation to hyperparameters, as it ensures that the parameter values remain within a fixed range, leading to more robust and stable performance. Figure 5 shows how the number of input connections for each node is normalized in the individual depicted in Figure 4 using Equation 1. After extracting the number of input connections for all nodes, these values are normalized to a range between 0 and 1.

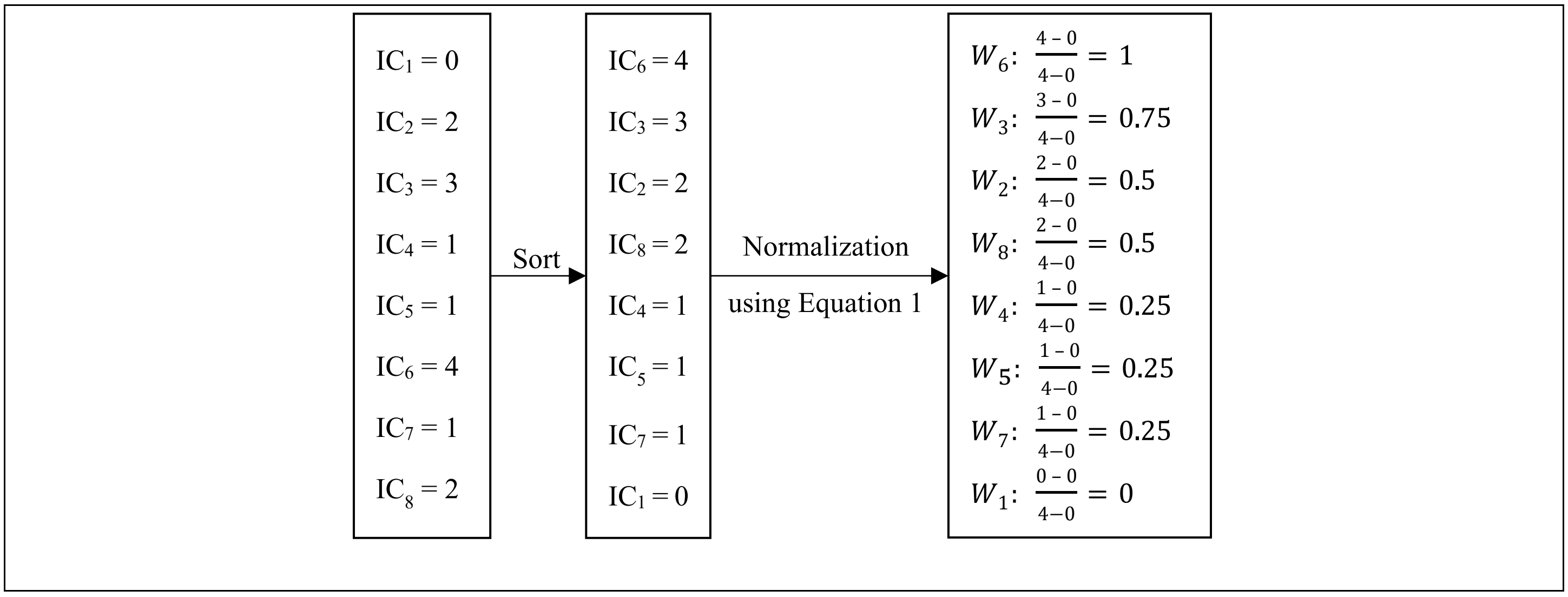

*Figure 5. Normalizing the number of input connections of the nodes of the individual shown in Figure 4.*

By normalizing the number of input connections, the importance value of each node is calculated. Nodes with a higher value are expected to have more influence on the individual and should have a lower $P_c$ (i.e., probability of crossover). In the standard GNP, two parents exchange the connections of the corresponding nodes with probability $P_c$, which is equal for all nodes and in all iterations. However, in the proposed method, the value of $P_c$ depends on the importance value of the node and the iteration number. Suppose we want to perform a crossover between individuals *i* and *j*. We must consider the *W* of both selected individuals *i* and *j* when calculating $P_c(n)$ for the corresponding node *n*, as shown in Equation 2.

$$P_c(n) = \Big(1 - t \times \big(W_i(n) + W_j(n) - 1\big)\Big) \times P_c \qquad t, W_i(n), W_j(n), P_c \in [0, 1] \tag{2}$$

where $W_i(n)$ and $W_j(n)$ are the importance values of node *n* in individuals *i* and *j*, respectively. $P_c$ is a constant value between 0 and 1, which equals the probability of crossover in standard GNP, and *t* is the iteration number normalized to the range [0, *UB*]. *UB* is the upper bound set by the system designer and is less than or equal to 1. The difference between nodes is meaningless at the early iterations because convergence has not yet begun. A node can have many input connections but may not contribute to achieving a high fitness score. In other words, exploitation cannot occur in the early iterations.

However, when the algorithm approaches the last iterations, the importance of exploitation increases. Therefore, in addition to the importance of nodes, $P_c(n)$ also depends on the iteration number. The rate at which $P_c(n)$ changes is an important hyperparameter that impacts the exploitation rate which is adjusted by *UB*. The value of *UB* determines how much $P_c(n)$ should be affected by *t*. If the algorithm should be affected more by this approach, the *UB* must be shifted toward 1, whereas when the *UB* is 0, the proposed crossover operates just like the standard crossover in GNP. For example, if we set the *UB* to 1, in the last iteration, the crossover probability $P_c(n)$ becomes 0 for the most important node of the two selected individuals. The diagrams in Figure 6 show the changing rate of $P_c(n)$ relative to $W_i(n) + W_j(n)$ and *t* when *UB* is set to 1 and 0.5, respectively. When *t* is set to 0, crossover works like the standard GNP. Gradually, as the iteration number increases, the difference between more important and less important nodes becomes greater.

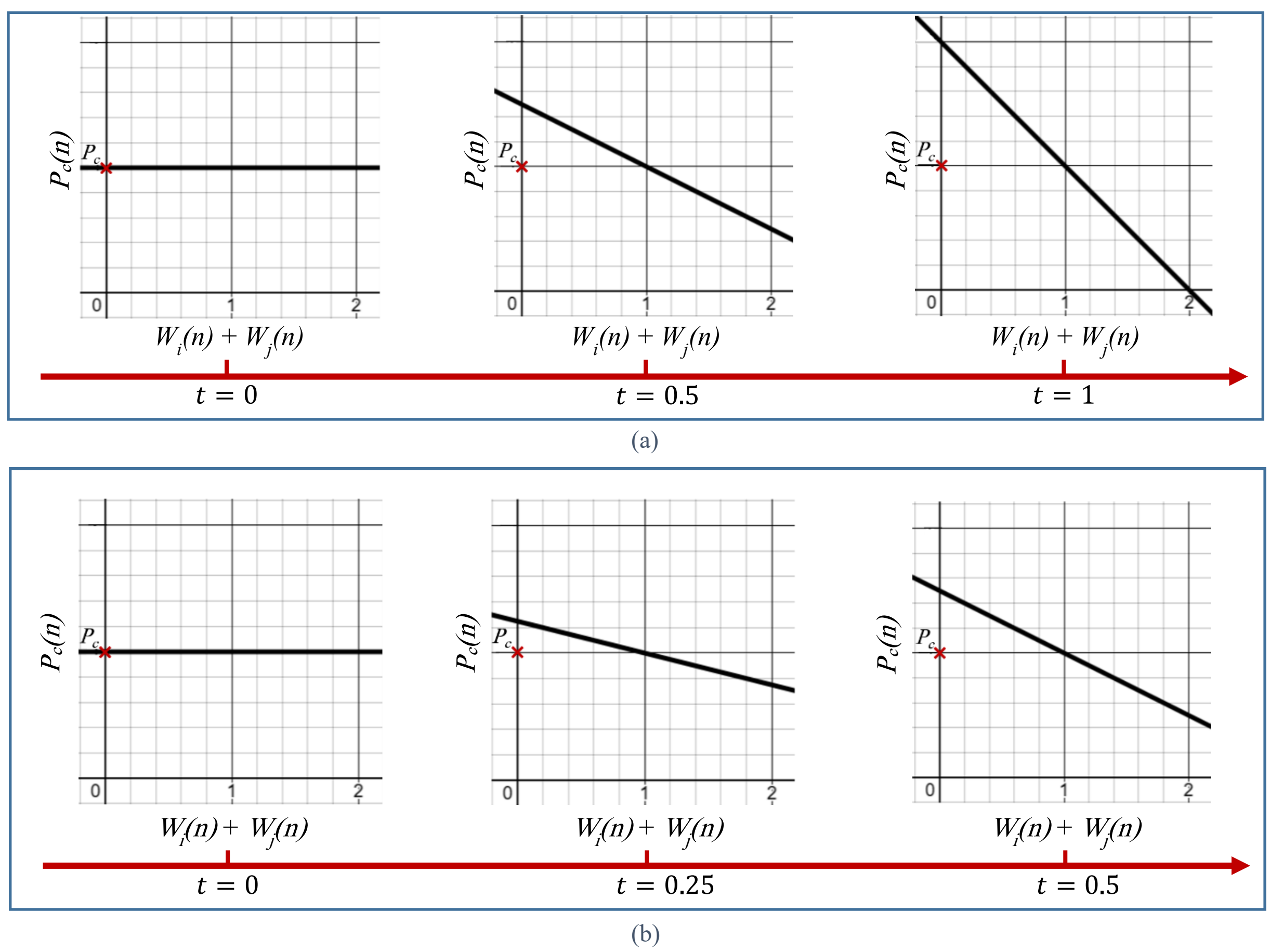

*Figure 6. The diagram of the $P_c(n)$ variations relative to the Wi(n) + Wj(n) and t when the UB is set to a) 1 and b) 0.5.*

In Equation 2, if the *UB* is set to any value between 0 and 1, the expected value of $P_c(n)$ will be $P_c$, as shown in Equation 3. This makes HGNP comparable to standard GNP by setting the same $P_c$ value. It demonstrates that the improvement of this algorithm is not due to a change in the average crossover probability, but rather to its adaptability.

$$E\big(P_c(n)\big) = \int_0^{UB} \left( \int_0^2 P_c(n)\, d\left(W_i(n) + W_j(n)\right)\right) dt = P_c \quad (3)$$

## 4.2. Exploration Phase

Using GNP, an agent can select the optimal action based on the environmental situation. This decision-making process can be compared to human behavior, with transitioning between judgment nodes analogous to thinking or deliberation and processing nodes analogous to action. We can enhance the algorithm's performance by drawing inspiration from human development across the lifespan and comparing behavioral patterns in early years with those in adulthood. In general, children tend to take more risks than adults [62]. They are eager to explore and learn about the world around them. They are more focused on taking action rather than thinking about the consequences. However, over time and through observing the feedback from their actions, they learn what is good or bad and how to behave more cautiously. They learn to be more careful and to think before acting [63, 64]. For instance, by touching a hot object, they learn it is hot and should not be touched. This concept of exploratory tendency in children can be applied to GNP by encouraging the agent to use more processing nodes instead of transitioning between judgment nodes during the early iterations.

In GNP, only the processing nodes navigate the agent towards the goal. The judgment nodes merely assist the agent in selecting the optimal processing nodes. Therefore, while the agent could potentially approach the goal by randomly selecting processing nodes, it is not feasible to achieve this by randomly utilizing judgment nodes and transitioning between them. This randomness in node selection is unavoidable during the early iterations. Consequently, in the early iterations, improvement is more probable through exploration via processing nodes rather than judgment nodes, especially in the GNP graphs that have more judgment nodes than processing nodes.

Mutation is a critical mechanism that supports exploration in evolutionary algorithms by introducing variability and allowing the algorithm to escape local optima. The degree of exploration driven by mutation must be carefully balanced to ensure efficient convergence toward global optima. This balance is typically achieved through parameter tuning or adaptive strategies, which adjust the mutation rate based on the algorithm's progress. Another mechanism that leads to improvement in the exploration is cycle elimination, which prevents the generation of cycles in the individual graph structure. This section improves the exploration mechanism by applying mutation modification and adding cycle elimination to the GNP algorithm.

### 4.2.1.Mutation

As mentioned before, at the start of running the algorithm, complex individuals that utilize a long sequence of judgment nodes are unlikely to have suitable fitness. Just as a young child, without sufficient experience, is unlikely to think about their actions deeply. Instead, we require exploration throughout the environment at the beginning, which can be achieved by prioritizing processing node selection over transitioning between judgment nodes. We need a mechanism to increase exploration and prevent the agent from overthinking. In the standard GNP, mutation is applied to an individual by selecting node connections with a constant mutation rate $P_m$ and randomly changing them to generate offspring. However, in our proposed method, in addition to standard mutation, we use a novel mutation operation to reduce the number of judgment nodes that point to other judgment nodes with an adaptive probability $P_{mNew}(n)$ for node $n$, which is adjusted according to the iteration number. An example of applying this mechanism is shown in Figure 7. This approach enhances the agent's ability to explore the environment. Indeed, this modified mutation operator changes the connection between judgment nodes with the rate of $P_{mNew}(n)$, which is proportional to the inverse of the iteration number. As the iteration number increases, the $P_{mNew}(n)$ decreases because exploration is needed more in the early iterations. Deep thinking or transitioning between judgment nodes becomes more rational as the population evolves in the last iterations. Equations 4 and 5 show that $P_{mNew}(n)$ can be calculated linearly or non-linearly based on the problem environment.

$$P_{mNew}(n) = (-\alpha_m * (t - 0.5) + 0.5) \times P_{mNew} \qquad \text{(Linear function)} \qquad (4)$$

$$P_{mNew}(n) = \frac{\alpha_m}{t + \alpha_m} \times P_{mNew} \qquad \text{(Non-linear function)} \qquad (5)$$

where $\alpha_m$ is the changing rate. $P_{mNew}$ is the parameter that indicates the average probability of the novel mutation. $t$ is the iteration number normalized between 0 and 1.

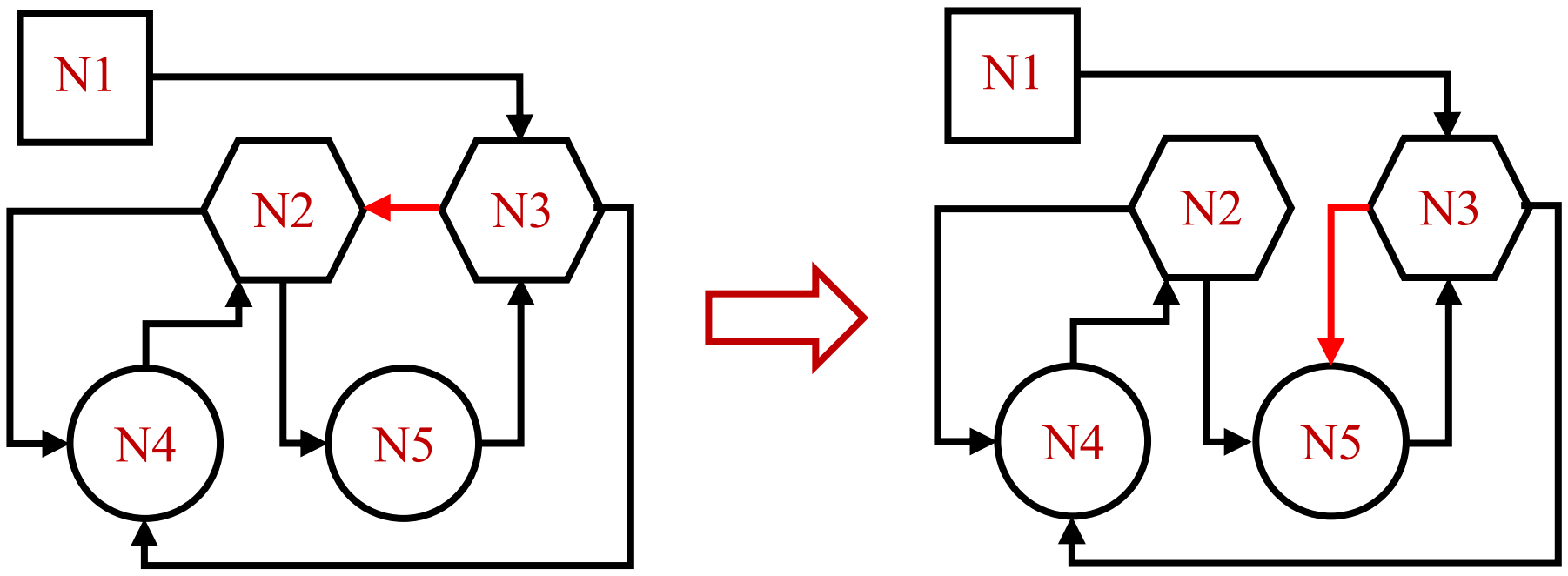

*Figure 7. This mechanism mutates the connections between judgment nodes.*

### 4.2.2.Cycle Elimination

One scenario that sometimes arises during evolution is getting stuck between nodes, resulting in an infinite loop. These loops can occur at different lengths depending on the number of nodes involved in their creation, as depicted in Figure 8. In the case of a cycle between processing nodes, the loop is infinite since it does not depend on the environmental situation. However, a change in the environmental state may break the loop when the cycle occurs between judgment nodes. In other words, the agent may be in a state of wise patience, waiting for other agents to act. However, this level of intelligence is unlikely to emerge in the early iterations. Furthermore, these cycles result in transitioning between judgment nodes, which delays the agent's exploration, as discussed in the previous section. In general, in the case of judgment nodes, only cycles with length one can be recognized by investigating the connection genes of a node. To find the cycles with a length greater than one, we need to access the sequence of actions chosen in the environment to identify which connections cause loops between them.

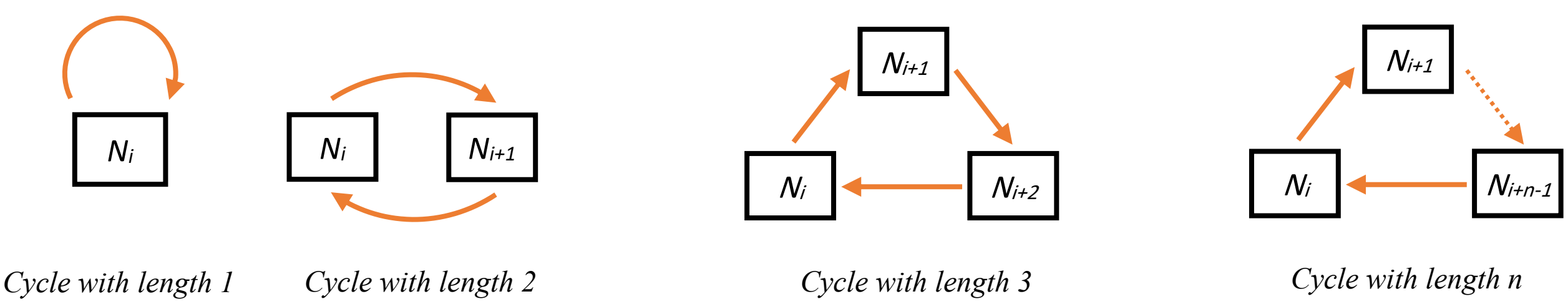

*Figure 8. The structure of loops varies based on the number of contributing nodes.*

In this research, two types of cycle eliminators were used for all individuals at each iteration. We employ a cycle eliminator with a probability of 1 for processing nodes, ensuring it always occurs. As the number of nodes contributing to the cycle increases, finding them becomes more difficult. Depending on the problem space and available computational resources, we can utilize different levels of cycle elimination for processing nodes. An example of a cycle with a length of two between processing nodes is depicted in Figure 9. To eliminate the cycle, one of the connections contributing to it must be modified. In the judgment cycle, cycle elimination is performed by $P_{elim}(n)$, which adapts according to the iteration number, and only cycles with length one are considered. The linear and non-linear equations for computing the value of $P_{elim}(n)$ are outlined in Equations 6 and 7, respectively.

$$P_{elim}(n) = (-\alpha_{elim} * (t - 0.5) + 0.5) \times P_{elim} \quad \text{(Linear function)} \quad (6)$$

$$P_{elim}(n) = \frac{\alpha_{elim}}{t + \alpha_{elim}} \times P_{elim} \quad \text{(Non-linear function)} \quad (7)$$

where $\alpha_{elim}$ is the changing rate of $P_{elim}(n)$, *and* $t$ is the iteration number normalized between 0 and 1. $P_{elim}$ is the parameter that indicates the average probability of judgment cycle elimination.

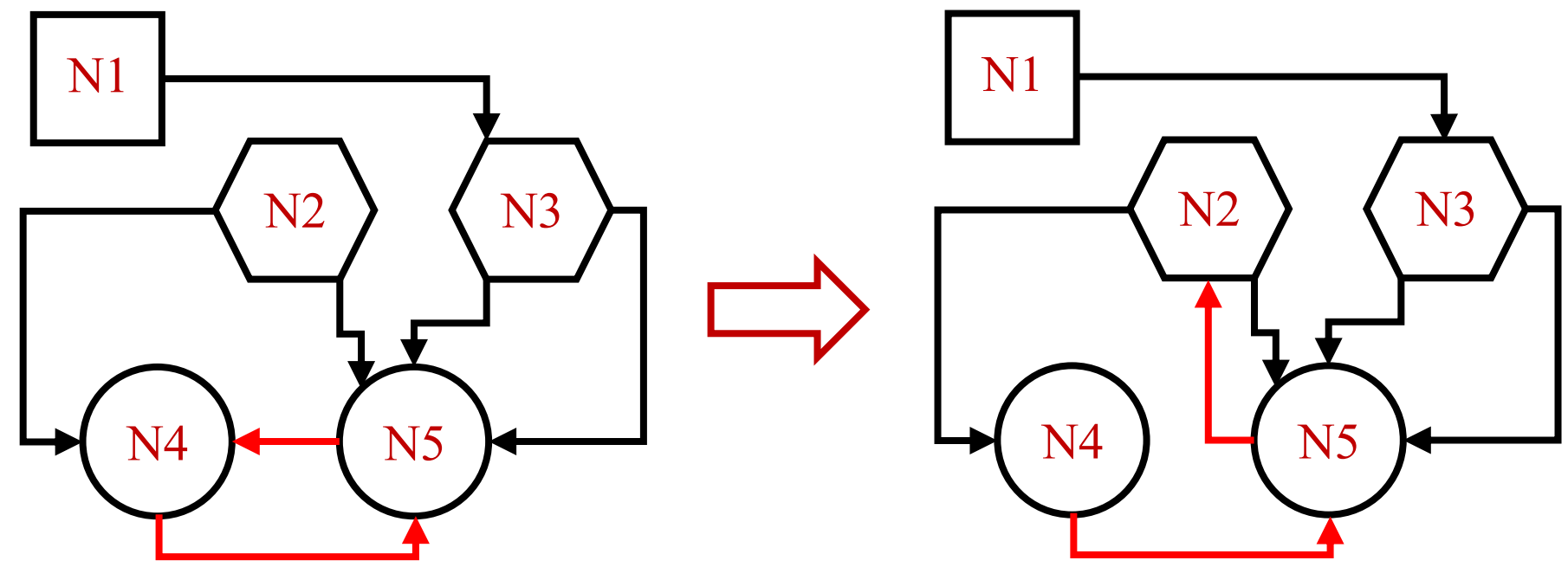

*Figure 9. An example of a cycle with length two for processing nodes.*

## 4.3. Integration of Exploitation and Exploration Phases

Using these methods, we can improve the evolutionary process in GNP by implementing more effective exploration-exploitation strategies. Furthermore, we have developed a framework to adjust the exploration-exploitation trade-off by tuning the methods' parameters. When more exploration is needed, we should increase the priority of action over deliberation, especially in early iterations. Conversely, when more exploitation is required, the focus shifts toward preserving influential nodes and enabling the agent to perform deeper reasoning. This results in better performance compared to tuning $P_c$ and $P_m$ in standard GNP by offering a more rational approach inspired by human developmental patterns. A pseudo-code of the HGNP algorithm is presented in Algorithm 1. It is important to note that during judgment cycle elimination, if we iterate over all nodes, $P_{elim}$ is calculated for each node involved in the cycle. This would result in $P_{elim} \times cycle_length$ instead of $P_{elim}$. To address this issue, $P_{elim}$ should be divided by the cycle length.

*Algorithm 1. HGNP pseudo-code*

1. **Initialize** population *Pop* consisting of the connections of graph structures.
2. While the termination condition is not fulfilled
    - 2.1. for ($i$=1: *PopulationSize*) // Selection
        - 2.1.1. **Calculate** fitness for *Pop*[$i$]: *fitnesses* [$i$].
    - 2.2. **Preserve** the elite individuals: $Pop_e$.
    - 2.3. for (i=1: *PopulationSize* / 2 – size($Pop_e$)) // Novel crossover
        - 2.3.1. **Select** two parents $P_1$ and $P_2$ from *Pop*.
        - 2.3.2. **Calculate** the number of input connections of each node ($n$) for $P_1$ and $P_2$: $IC_P$.
        - 2.3.3. **Normalize** $IC_P$ into the range [0, 1] indicating node importance: $W_P$.
        - 2.3.4. **Calculate** the probability of crossover for each node using $W_{P1}(n)$, $W_{P2}(n)$ and $t$: $P_c$.
        - 2.3.5. **Perform** crossover with the probability $P_c$.
    - 2.4. for ($i$=1: *populationSize* – size($Pop_e$)) // Mutation and cycle elimination
        - 2.4.1. **Perform** mutation on all connections of *Pop*[$i$] with $P_m$. // Traditional mutation
        - 2.4.2. **Perform** mutation on connections between judgment nodes of *Pop*[$i$] with $P_{mNew}$. // Novel mutation
        - 2.4.3. if (*Pop*[$i$] contains a processing cycle) // Processing cycle eliminator
            - 2.4.3.1. **Select** one of the nodes in the processing cycle with equal probability.
            - 2.4.3.2. **Modify** the selected node's connections to break the loop.
        - 2.4.4. if (*Pop*[$i$] contains a judgment cycle) // Judgment cycle eliminator
            - 2.4.4.1. if (random_seed < $P_{elim}$ / *cycleLength*)
                - 2.4.4.1.1. **Modify** the selected node's connections to break the loop.

This approach adopts a "child-to-adult" perspective that aligns with human developmental patterns, as illustrated in Figure 10. Children initially explore their environment through action (analogous to prioritizing processing nodes in HGNP) and reaction (fitness evaluation) during early ages (early iterations). Over time, this shifts toward a "think first, then act" behavior (transitions between judgment nodes), where decisions are increasingly guided by accumulated experience (exploitation of influential nodes) as individuals grow older (the iteration number increases).

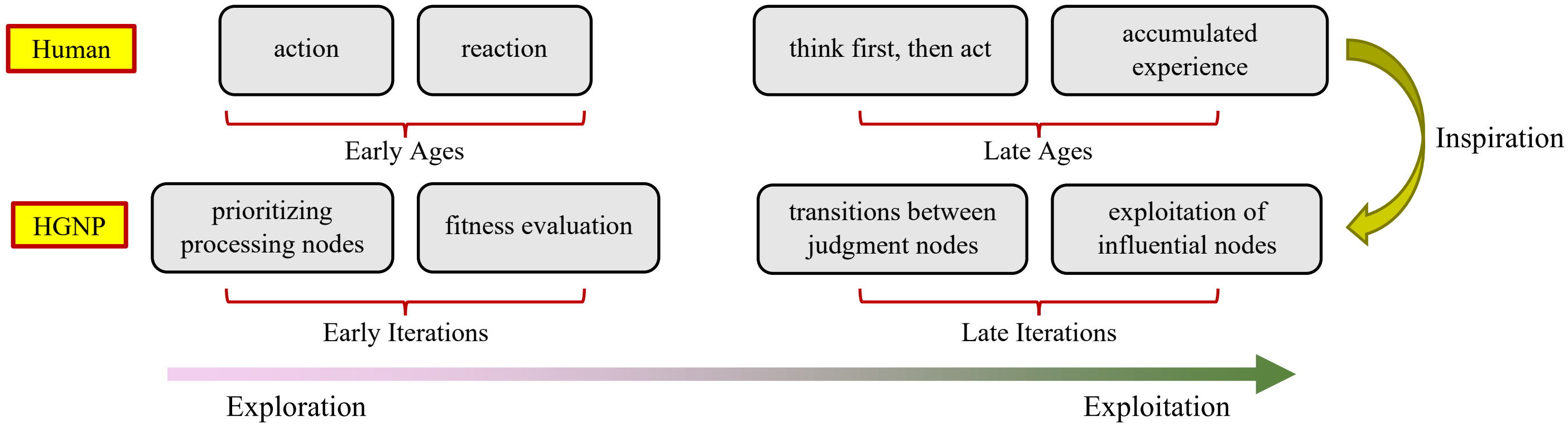

*Figure 10. Human developmental inspiration underlying the adaptive exploration–exploitation mechanism in HGNP.*

Our proposed algorithm can be applied to all GNP algorithms that use the standard crossover operation. Since the novel mutation and cycle elimination mechanisms are independent of other components, they can be applied to any graph structure that utilizes judgment and processing nodes. However, because the proposed crossover modifies the node selection process, it requires the presence of an existing crossover operator to function.

# 5. Experimental Results

To evaluate the effectiveness of the proposed method, we applied it to the Tileworld environment [65], a well-established benchmark for studying agent control problems. Tileworld is frequently used as a primary benchmark in GNP research due to its dynamic and challenging nature [40, 43, 61]. It provides a controlled, dynamic testbed for evaluating agents' performance in solving planning and decision-making problems. By applying our method to this testbed, we aimed to compare its performance with the standard GNP and two of its extensions. In this section, we briefly introduce the benchmark environment. Then, we compare the results of applying our proposed method to standard GNP and two extensions of this algorithm.

## 5.1. Tileworld

The Tileworld environment consists of a grid of cells on which various objects could exist, including agents, tiles, holes, obstacles, and floors. The 11 × 11 two-dimensional grid is used in this research. The goal is to push tiles into holes. Once a tile is pushed into a hole, it becomes part of the floor, and both the tile and the hole disappear. The agents can sense their surrounding environment and, based on the outputs of sensors, decide which action to take. These sensors are referred to as judgment nodes in the GNP graph. Notably, the environment is highly dynamic, as agents, tiles, and holes are continuously moved or removed during interaction, causing the state of the world to change over time. In our experiment, the sequence of sensing the environment using judgment nodes and taking action according to the output of the judgment nodes is accomplished by using seven judgment functions and four processing functions listed in Table 1. These settings are consistent with previous GNP studies [40, 45, 61]. Using the first four judgment functions, agents can sense what exists in their adjacent cells. The next three functions enable agents to sense the relative directions of tiles and holes in the environment. Based on the returned arguments, the agent can perform one of the following four movement actions: Move Forward, Turn Left, Turn Right, or Stay. By combining the various functions listed in Table 1, programs for controlling the agents can be generated. These programs are designed to guide the agents in pushing all tiles into the holes. The number of instances of each node in the graph is manually determined and is typically the same for all nodes. This number is called the program size. For example, if we set the program size to five, we have five instances of each judgment and processing node in each individual.

*Table 1. List of judgment and processing nodes used in the Tileworld environment.*

| Node type | Symbol | Description | outputs |
|---|---|---|---|
| Judgment nodes | WEF | **W**hat **E**xist **F**ront | Agent, Tile, Hole, Obstacle, Nothing |
| | WER | **W**hat **E**xist **R**ight | |
| | WEL | **W**hat **E**xist **L**eft | |
| | WEB | **W**hat **E**xist **B**ack | |
| | NTD | **N**earest **T**ile **D**irection | Forward, Backward, Right, Left |
| | SNTD | **S**econd **N**earest **T**ile **D**irection | |
| | NHD | **N**earest **H**ole **D**irection | |
| Processing nodes | MF | **M**ove **F**orward | --- |
| | TR | **T**urn **R**ight | |
| | TL | **T**urn **L**eft | |
| | ST | **ST**ay | |

To clarify how these functions operate, the outcomes of applying them under the conditions depicted in Figure 11 (a) are presented in Figure 11 (b). This example shows how forward, backward, left, and right directions are considered.

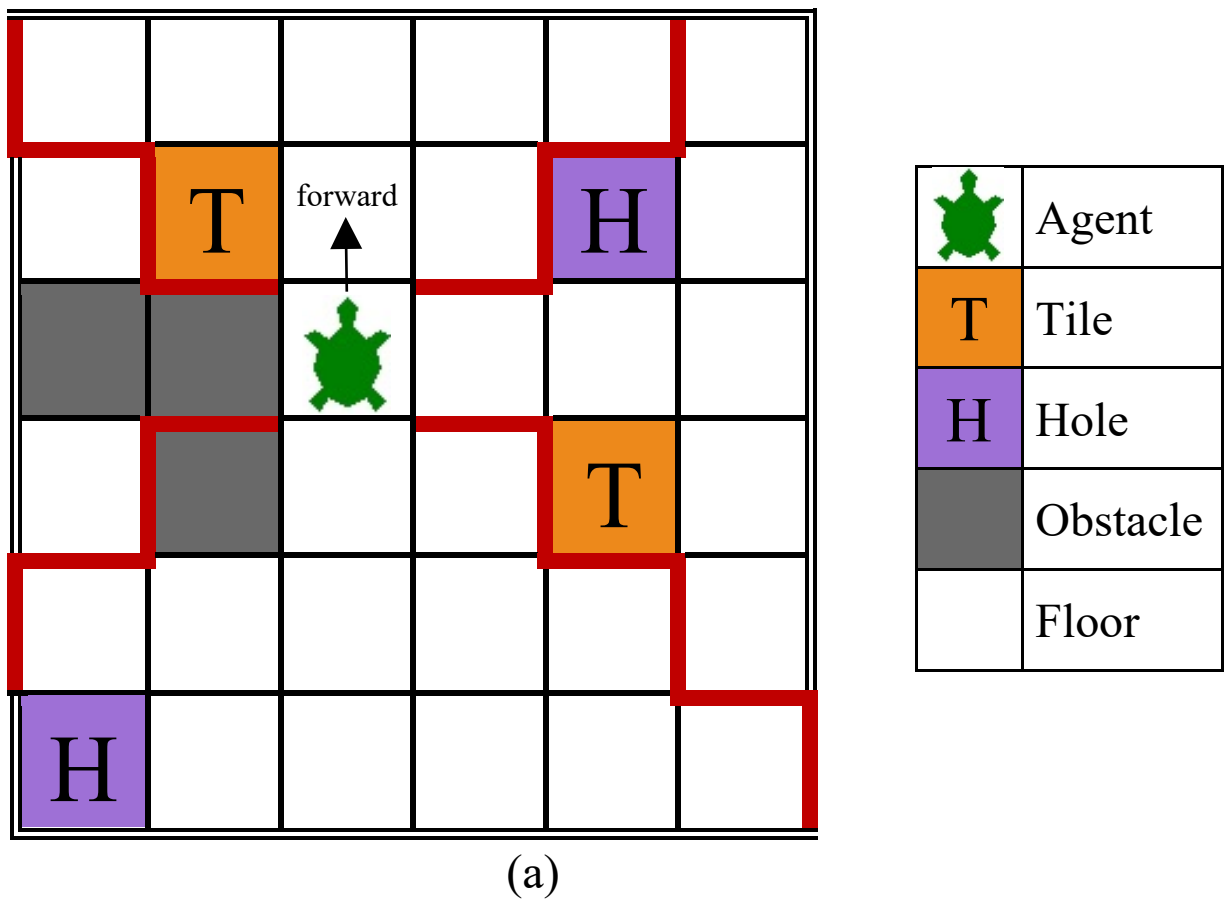

(a)

| Judgment functions | outputs |
|---|---|
| WEF | Nothing |
| WER | Nothing |
| WEL | Obstacle |
| WEB | Nothing |
| NTD | Forward |
| SNTD | Right |
| NHD | Right |

(b)

*Figure 11. An example of the outputs of judgment functions.*

Three factors are considered when assessing the behavior of the agents. Within a limited number of steps, the agents should 1) push as many tiles into holes as possible, 2) do so using the minimum number of steps, and 3) if any tiles remain, they should be moved as close as possible to the nearest holes. Consequently, the fitness function for the solutions in the Tileworld system is formulated as follows:

$$f = w_1 \times \underbrace{DT}_{1)} + w_2 \times \underbrace{(IS - TS)}_{2)} + w_3 \times \underbrace{\left[\sum_{t=1}^{T}\left(ID_{(t)} - FD_{(t)}\right)\right]}_{3)} \tag{8}$$

where $DT$ is the number of tiles dropped into the holes by the agents. $IS$ is the user-defined maximum number of steps that an agent can take. $TS$ is the number of steps taken. $T$ is the number of tiles in the environment. $ID_{(t)}$ represents the initial distance from tile $t$ to its nearest hole, and $FD_{(t)}$ represents the distance after $TS$ steps. $w_1$, $w_2$, and $w_3$ are the parameters controlling the weights of terms 1, 2, and 3. We set these parameters in our experiment to $w_1$=100, $w_2$=1, and $w_3$=20. These settings follow commonly used configurations in related studies.

## 5.2. Compared Algorithms and Experimental Settings

The proposed algorithm can be applied to any GNP variant that uses standard crossover. So, we applied HGNP to standard GNP [12] and two recently established algorithms, Simplified_GNP [61] and Situation-based GNP [45]. Simplified_GNP was proposed to address unnecessary evolutionary difficulties and issues with invalid/negative evolution by considering the "transition by necessity" feature of the directed graph. This method works by evolving only the branches that are activated during execution. Simplified crossover exclusively operates on branches transited in at least one parent individual. Simplified mutation is restricted to those branches that have been transited during execution. This reduces the search space by focusing only on transited branches. Situation-based GNP (SBGNP) is designed for environments with multiple agents. Rather than seeking a single optimal strategy for all agents, individual strategies are developed for each agent based on its specific situation. This approach allows the goal to be achieved more easily and quickly, as finding a unified strategy for all agents is more challenging than creating separate strategies suited to each agent's circumstances. However, the drawback of this model is the increased search space, as it requires finding separate strategies for each agent rather than a single strategy for all agents. The search spaces of GNP and SBGNP are shown in Equation 9 and Equation 10, respectively.

$$Search_Space_{\mathrm{GNP}} = (PS \times (NJ \times NB + NP))^{PS \times (NJ + NP)} \tag{9}$$

$$Search_Space_{\mathrm{SBGNP}} = Search_Space_{\mathrm{GNP}} \times NA \tag{10}$$

The number of judgment and processing functions is $NJ$ and $NP$, respectively. $PS$ shows the program size, and each judgment function has an average of $NB$ branches. $NA$ is the number of agents.

Table 2 details the settings used for all GNP variants, divided into: 1) shared parameters across all methods and 2) HGNP parameters, which were adjusted for HGNP and the methods integrated with it. The hyperparameters were selected experimentally based on iterative testing to optimize the methods' performance. The non-linear function was used for the novel mutation and cycle elimination stages. All experiments were conducted on a laptop with an Intel® Core™ i7-6700HQ

CPU and 16 GB of RAM. The operating system was Windows 10, and the programs were developed in Python. For visualization purposes, the Turtle and Tkinter libraries were used.

*Table 2. Parameter Settings for the Tileworld System*

| Shared parameters | | |
|---|---|---|
| | Population Size | 100 |
| | Elite individuals | 2 |
| | Crossover rate ($P_c$) | 0.4 |
| | Traditional mutation rate ($P_m$) | 0.01 |
| | Initial Steps (*IS*) | 60 |
| | Maximum number of iteration | 1000 |
| HGNP parameters | | |
| | Novel mutation rate ($P_{mNew}$) | 0.15 |
| | $\alpha_m$ | 0.15 |
| | Cycle elimination rate ($P_{elim}$) | 0.8 |
| | $\alpha_{elim}$ | 0.5 |
| | Judgment nodes cycle length elimination | 1 |
| | Processing nodes cycle length elimination | 1 and 2 |
| | *UB* | 1 |

To ensure a fair comparison, we conducted experiments in three environments to evaluate the performance of each algorithm more comprehensively. These environments are illustrated in Figure 12 and arranged in order of increasing difficulty. In these scenarios, the number of agents, tiles, and holes is set to 3, but with different positions. The proposed method was tested with two program sizes of 3 and 5, and on three environments, as shown in Figure 12.

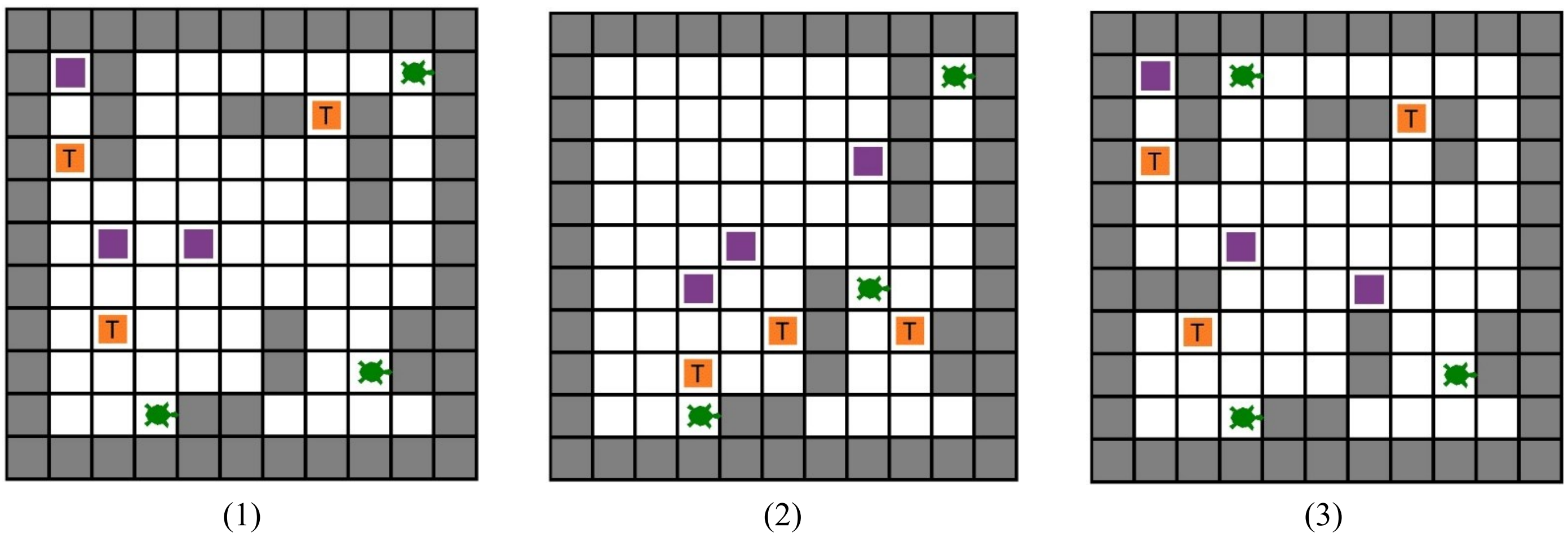

*Figure 12. Tileworld environments used in the experiment, in order of increasing difficulty.*

For a more reliable comparison, each algorithm was run 50 times. Figure 13 illustrates the performance of the methods, showing the mean of fitness of 50 executions. *PS* and *env* stand for program size and environment number, respectively. We colorized the fitness curve of each version with a darker color when combined with our proposed method for better comparison. For example, simplified_GNP is shown in light green, while HGNP-simplifiedGNP is in dark green.

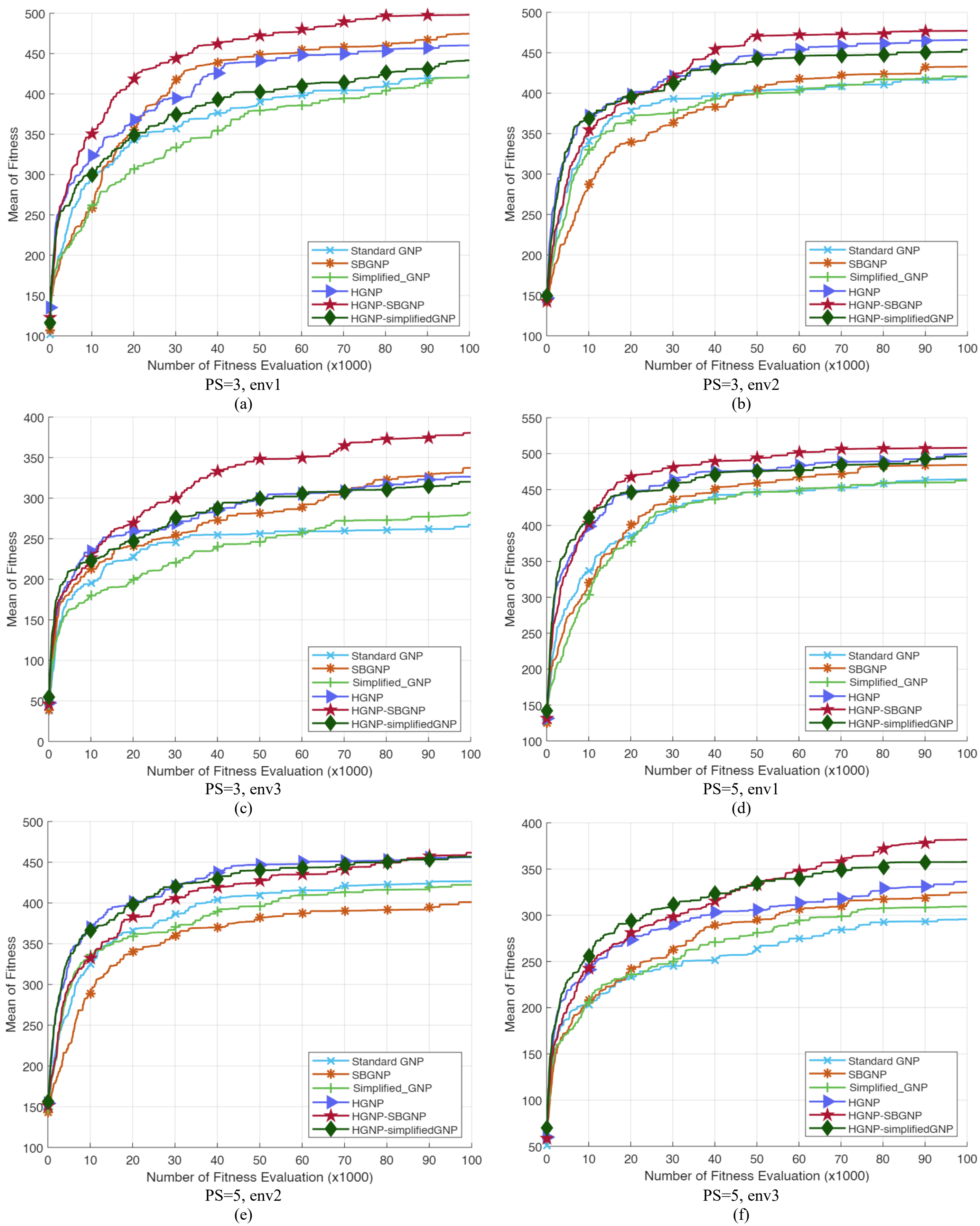

*Figure 13. fitness curve of the best solutions in different environments and program sizes*

It is clear that HGNP-SBGNP, which combines HGNP with SBGNP, achieves the best fitness. As reported in [45], SBGNP generally produces better results than standard GNP when using a smaller program size. Similar results are observed in these tests as better performance is achieved when PS is set to three. SBGNP has a larger search space because it has a separate graph for each agent. This larger space enhances HGNP's exploration potential when integrated with SBGNP. Additionally, this explains why SBGNP has lower fitness than other algorithms in the initial iterations. Finding appropriate solutions requires more iterations in this larger search space. For example, in Figure 13 (b), this method had the worst performance until the middle of the iterations but eventually outperformed GNP and simplified_GNP. A similar trend was observed with HGNP-SBGNP, which initially showed worse results than HGNP and HGNP-simplifiedGNP but ultimately achieved the highest fitness. Simplified_GNP has a smaller search space, which is its strength. However, in this case, it limits the potential for improvement when incorporated with HGNP. This is why HGNP-simplifiedGNP has lower fitness than HGNP in Figure 13 (a) and (b).

HGNP improves the performance of methods compared to their base versions in all cases. Also, this improvement can be achieved with fewer iterations, thanks to the enhanced exploration ability provided by the novel mutation and cycle elimination proposed in this research. Notably, no improvement was observed when we tested only the novel crossover approach; in some cases, performance even worsened. This decline occurred due to the nature of agentic systems and GNP benchmarks, such as Tileworld, which require a high level of exploration. If increased exploitation leads to decreased exploration, overall performance suffers. On the other hand, when we used only the two exploration methods, performance was good in the early iterations; however, after that point, no significant improvement was observed compared to when all three proposed modifications were used together. However, its overall performance was better than standard GNP in all situations. Consequently, the two exploration methods lead to better results in early iterations, while the proposed crossover helps retain the experiences gained during runtime.

The detailed statistical analysis of the results is presented in Table 3-8. For each column, the best performance is highlighted in bold. Additionally, p-values of the t-test [66] less than 0.05 are bolded. When comparing sample sets from two different algorithms, the p-value indicates the significance of the difference between them. A p-value less than 0.05 shows that one algorithm outperforms the other with a confidence level greater than 95%, indicating a statistically significant difference. We added two columns for p-values. The first column is for the p-value of the best method relative to the others, and the second column is the p-value of each method relative to its integration with HGNP to demonstrate the improvement achieved by applying HGNP to them. All algorithms were tested 50 times. The number of successful solutions points to the solutions where all tiles are dropped into the holes successfully.

As shown in Table 3-5, when the program size was set to 3, HGNP performed better than baseline methods (non-HGNP-integrated methods) in environment 2. SBGNP ranked second in environments 1 and 3. Compared to other baseline methods, it demonstrated the best performance across all three environments. When the program size was increased to five, its performance declined. HGNP-SBGNP achieved the best results in all experiments and had the lowest standard deviation in Environment 1 for both program sizes. When the program size is set to five, it achieved almost perfect results with a standard deviation of 3.44 and produced 50 successful solutions. This suggests that selecting a sufficiently large program size for the given environmental conditions allows HGNP-SBGNP to represent even stronger performance.

HGNP outperforms the three basic methods when the program size is five, showing a significant improvement in average fitness, as detailed in Table 6-8. Additionally, HGNP consistently produced p-values lower than 0.05 in all scenarios. In environments 1 and 2, HGNP-integrated algorithms achieved better standard deviations than the basic methods, making these algorithms more reliable. However, this was not the case in environment 3, where program sizes of three or five are insufficient for the difficulty of the environment. Simplified_GNP had the worst performance in terms of standard deviation, although it achieved better results than standard GNP in average fitness. When this algorithm was integrated with HGNP, its average fitness and standard deviation improved.

*Table 3. Results of algorithms in environment 1 when the program size is 3.*

| Algorithm Name | Rank | Average | Standard Deviation | Number of Successful Solutions | p-value | p-value Relative to Its HGNP-Integrated Version | Max of fitness | Min of fitness |
|---|---|---|---|---|---|---|---|---|
| GNP | 6 | 421.12 | 88.69 | 28 | **5.41097E-08** | **0.015604** | 508 | 280 |

| simplified_GNP | 5 | 423.12 | 109.44 | 33 | **1.5538E-05** | 0.187957 | **512** | 140 |
|---|---|---|---|---|---|---|---|---|
| SBGNP | 2 | 474.66 | 69.66 | 42 | **0.023090884** | **0.023091** | **512** | 280 |
| HGNP | 3 | 460.14 | 76.02 | 39 | **0.001889675** | | **512** | 280 |
| HGNP-simplifiedGNP | 4 | 441.52 | 86.22 | 35 | **6.1344E-05** | | **512** | 205 |
| **HGNP-SBGNP** | **1** | **498.28** | **34.51** | **48** | | | **512** | **325** |

*Table 4. Results of algorithms in environment 2 when the program size is 3.*

| Algorithm Name | Rank | Average | Standard Deviation | Number of Successful Solutions | p-value | p-value Relative to Its HGNP-Integrated Version | Max of fitness | Min of fitness |
|---|---|---|---|---|---|---|---|---|
| GNP | 6 | 420.12 | 79.33 | 26 | **4.96483E-05** | **0.002387** | 509 | 285 |
| simplified_GNP | 5 | 420.92 | 83.41 | 28 | **5.58156E-05** | **0.010918** | 507 | 202 |
| SBGNP | 4 | 432.72 | 77.25 | 30 | **0.000276195** | **0.000276** | **511** | 285 |
| HGNP | 2 | 465.66 | 63.60 | 41 | 0.160185337 | | 507 | 285 |
| HGNP-simplifiedGNP | 3 | 453.78 | 67.37 | 37 | 0.016453541 | | 505 | 322 |
| **HGNP-SBGNP** | **1** | **477.08** | **48.69** | **44** | | | 510 | **347** |

*Table 5. Results of algorithms in environment 3 when the program size is 3.*

| Algorithm Name | Rank | Average | Standard Deviation | Number of Successful Solutions | p-value | p-value Relative to Its HGNP-Integrated Version | Max of fitness | Min of fitness |
|---|---|---|---|---|---|---|---|---|
| GNP | 6 | 267.22 | 80.34 | 2 | **1.91783E-08** | **0.000964** | 498 | 156 |
| simplified_GNP | 5 | 282.16 | 93.60 | 4 | **7.6275E-07** | **0.036668** | 488 | 62 |
| SBGNP | 2 | 337.44 | **84.05** | 9 | **0.011023245** | **0.011023** | 497 | 194 |
| HGNP | 3 | 326.42 | 88.49 | 9 | **0.0033987** | | 494 | 156 |
| HGNP-simplifiedGNP | 4 | 320.24 | 84.38 | 7 | **0.001324826** | | 496 | 156 |
| **HGNP-SBGNP** | **1** | **380.46** | 88.45 | **18** | | | **501** | **199** |

*Table 6. Results of algorithms in environment 1 when the program size is 5.*

| Algorithm Name | Rank | Average | Standard Deviation | Number of Successful Solutions | p-value | p-value Relative to Its HGNP-Integrated Version | Max of fitness | Min of fitness |
|---|---|---|---|---|---|---|---|---|
| GNP | 5 | 464.42 | 75.05 | 39 | **7.97736E-05** | **0.001729** | **512** | 302 |
| simplified_GNP | 6 | 462.52 | 74.29 | 39 | **4.5066E-05** | **0.003823** | **512** | 293 |
| SBGNP | 4 | 484.28 | 60.70 | 44 | **0.003877105** | **0.003877** | **512** | 302 |
| HGNP | 2 | 499.8 | 25.93 | 49 | **0.014377474** | | **512** | 325 |
| HGNP-simplifiedGNP | 3 | 496.18 | 35.34 | 48 | **0.011851657** | | **512** | 325 |
| **HGNP-SBGNP** | **1** | **508.14** | **3.44** | **50** | | | **512** | **499** |

*Table 7. Results of algorithms in environment 2 when the program size is 5.*

| Algorithm Name | Rank | Average | Standard Deviation | Number of Successful Solutions | p-value | p-value Relative to Its HGNP-Integrated Version | Max of fitness | Min of fitness |
|---|---|---|---|---|---|---|---|---|
| GNP | 4 | 426.66 | 79.44 | 29 | **0.007271271** | **0.031372311** | 507 | 285 |
| simplified_GNP | 5 | 422.36 | 78.79 | 28 | **0.005446218** | **0.014863412** | 507 | 285 |
| SBGNP | 6 | 401.46 | 76.22 | 20 | **2.01778E-05** | **2.01778E-05** | 505 | 285 |
| HGNP | 3 | 456.04 | 68.36 | 37 | 0.348101972 | | **509** | **322** |
| HGNP-simplifiedGNP | 2 | 456.78 | 65.77 | 38 | 0.354881793 | | 507 | **322** |
| **HGNP-SBGNP** | **1** | **461.66** | **62.85** | **39** | | | **509** | **322** |

*Table 8. Results of algorithms in environment 3 when the program size is 5.*

| Algorithm Name | Rank | Average | Standard Deviation | Number of Successful Solutions | p-value | p-value Relative to Its HGNP-Integrated Version | Max of fitness | Min of fitness |
|---|---|---|---|---|---|---|---|---|
| GNP | 6 | 295.76 | **77.35** | 3 | **1.93421E-05** | **0.010784** | 495 | 156 |
| simplified_GNP | 5 | 309.46 | 97.49 | 7 | **0.000271139** | **0.002965** | 497 | 140 |
| SBGNP | 4 | 324.66 | 86.95 | 7 | **0.001999093** | **0.001999** | **507** | 182 |
| HGNP | 3 | 336.56 | 90.79 | 10 | **0.004379768** | | 499 | 194 |
| HGNP-simplifiedGNP | 2 | 357.6 | 85.73 | 13 | 0.110145915 | | 501 | 194 |
| **HGNP-SBGNP** | **1** | **381.86** | 97.08 | **20** | | | 505 | **199** |

Table 9 presents the computational times of the compared algorithms. The results indicate no significant differences in computational time among these methods, demonstrating that all algorithms operate within a comparable efficiency range. Notably, the proposed method maintains this consistency, as it does not introduce any substantial overhead or increase in computational time compared to others. This ensures that the improvements achieved in performance do not come at the cost of run time.

*Table 9. The average computational time of algorithms in seconds*

| **Algorithm Name** | **Computation Time (s)** |
|---|---|
| GNP | 425.4215 |
| simplified_GNP | 557.8376 |
| SBGNP | 579.9667 |
| HGNP | 521.7812 |
| HGNP-simplifiedGNP | 469.1345 |
| HGNP-SBGNP | 567.3635 |

To further illustrate the algorithm's behavior, three videos demonstrating HGNP's execution in environments 1-3 are provided in the supplementary materials.

## 5.3. Discussion

We proposed three improvements to the GNP algorithm inspired by human decision-making behavior. These improvements result in better convergence by using enhanced exploitation and exploration mechanisms. To improve the exploitation phase, we applied a modified crossover approach to preserve influential nodes in the graph structure. For exploration, we introduced a novel mutation that reduces the number of connections between judgment nodes, increasing the chance of exploring the environment in the early iterations. Another approach to improve exploration was the elimination of processing and judgment cycles in GNP graphs, which helps the algorithm avoid getting stuck in loops, which would reduce the exploration potential. The experimental results demonstrate that HGNP outperformed the other compared algorithms, both 1) as a standalone method (i.e., when integrated with standard GNP), and 2) when integrated with other algorithms.

In comparison with adaptive GNP [42], while both methods dynamically adjust crossover and mutation probabilities, HGNP introduces a novel perspective on the exploration and exploitation process. Rather than relying solely on the traditional view of crossover as exploitation and mutation as exploration, HGNP defines transitions between judgment nodes as exploitation and transitions between processing nodes as exploration. This paradigm is new among GNP and its variants, allowing for more precise control over the exploration–exploitation trade-off throughout the evolutionary process. In Adaptive GNP, operator probabilities are modified to bias the search toward preserving the frequently reused nodes in high-quality individuals. In contrast, inspired by human behavioral patterns, HGNP adapts the parameters of these analogized operators based on the normalized iteration number. This method is inspired by a "child-to-adult" perspective, where children tend to explore through an action-reaction mechanism, whereas adults exploit their accumulated experience by thinking and taking more conservative steps. Furthermore, in certain scenarios, a node with minimal reuse can still play a crucial role in the agent's strategy. Such a node would be susceptible to disruption under frequency-based updating (as performed in adaptive GNP), whereas adapting parameters based on iteration number fosters a more reliable evolutionary process overall.

This approach offers several notable advantages. It demonstrates improved evolutionary convergence, particularly in GNPs with larger search spaces, such as SBGNP, without introducing significant computational overhead. Additionally, HGNP provides a systematic framework to balance exploration and exploitation effectively. The exploration strategies can be applied to all GNP algorithms, and the exploitation strategy is compatible with all GNPs that use standard crossover, which makes HGNP adaptable to diverse contexts.

Despite its strengths, HGNP also has some limitations. One of the main challenges is its sensitivity to hyperparameters, meaning that fine-tuning is required for optimal performance. Furthermore, alternative functions for calculating $P_c$, $P_{mNew}$, and $P_{elim}$ may offer better convergence in certain applications. However, this framework opens new avenues for developing novel methods for hyperparameter tuning in the future. HGNP also shows limited improvement when applied to algorithms operating within smaller search spaces, such as Simplified_GNP, as its restricted search space limits HGNP's potential to enhance performance.

The proposed HGNP method is broadly applicable to agentic systems that require adaptive decision-making and efficient search strategies. In dynamic and uncertain environments, such as robotics, HGNP can enhance agents' ability to explore new strategies while maintaining the effective exploitation of learned behaviors. Additionally, HGNP's capability to balance exploration and exploitation makes it suitable for applications in automated trading, where adaptive strategies are crucial for responding to market fluctuations. The method can also contribute to Explainable AI (XAI) by providing interpretable decision-making structures in AI-driven systems. Furthermore, HGNP's ability to enhance exploration capability can benefit graph-based LLM agents, where efficient exploration within their reasoning processes is essential [67].

# 6. Conclusion

In this study, we proposed HGNP, a framework featuring three key modifications—adaptive crossover, adaptive mutation, and cycle elimination. These innovations address a critical limitation in traditional GNP methods: the lack of an adaptive framework to balance exploration and exploitation in self-evolving agentic systems. By transitioning from action-oriented exploration in early iterations to more cautious strategies in later iterations, HGNP provides a more efficient evolutionary path. This is achieved by analogizing transitions between judgment nodes as thinking and processing nodes as action. Our experiments across various environments and program sizes demonstrated that using HGNP techniques integrated with standard GNP and its two variants, i.e., simplified_GNP and SBGNP, improves their final solutions, fitness, and standard deviation. One of the important strengths of HGNP is its compatibility with all GNP variants that utilize standard crossover operations. Testing this compatibility, we applied it and observed significant improvements in SBGNP and notable improvements in simplified_GNP.

Although this study demonstrates the effectiveness of HGNP in enhancing GNP, various opportunities for future research remain uninvestigated. Integrating RL mechanisms can enable real-time dynamic adjustment of the exploration–exploitation balance based on environmental feedback, rather than relying on predefined functions. This enhancement could further improve HGNP's adaptability in non-stationary and highly dynamic environments, benefiting real-world complex scenarios such as financial markets, robotics, and personalized recommendation systems. Additionally, the current HGNP model is primarily designed for single-objective problems. Future research can focus on extending HGNP to Multi-Objective Genetic Network Programming (MOGNP), where balancing exploration and exploitation involves managing trade-offs among multiple conflicting objectives, such as accuracy, interpretability, and computational cost. Finally, since GNP has inherent explainability due to its graph structure, future work could involve designing an interactive visualization tool that allows human users to interpret, manipulate, and optimize HGNP-generated graphs easily. This aligns with Industry 5.0 goals of human-AI collaboration and could improve the trust and usability of HGNP-based systems.

# Code Availability

The source code used in this study is publicly available at: https://github.com/alikohan/gnp.

# CRediT authorship contribution statement

**Ali Kohan:** Conceptualization, Methodology, Software, Validation, Investigation, Data curation, Writing – original draft, Writing – review and editing, Visualization. **Mohamad Roshanzamir:** Validation, Investigation, Writing – original draft, Writing – review and editing, Supervision, Project administration. **Roohallah Alizadehsani:** Writing – review and editing. **Seyedali Mirjalili:** Validation, Writing – review and editing, Supervision.

# References

[1] S. Murugesan, "The rise of agentic AI: implications, concerns, and the path forward," *IEEE Intelligent Systems,* vol. 40, no. 2, pp. 8-14, 2025.

[2] H.-a. Gao *et al.*, "A survey of self-evolving agents: On path to artificial super intelligence," *arXiv preprint arXiv:2507.21046,* vol. 1, 2025.

[3] T. Wei *et al.*, "Agentic reasoning for large language models," *arXiv preprint arXiv:2601.12538,* 2026.

[4] L. Longo *et al.*, "Explainable Artificial Intelligence (XAI) 2.0: A manifesto of open challenges and interdisciplinary research directions," *Information Fusion,* vol. 106, p. 102301, 2024.

[5] A. Kohan *et al.*, "Human-Centered Pathways to Trustworthy AI in Healthcare: A Comparative Analysis of Explainable AI, Human-in-the-Loop, Hybrid AI, and Uncertainty Quantification Techniques," 2026.

[6] A. Kohan, A. Zahedi, R. Alizadehsani, R. S. Tan, and U. R. Acharya, "Application of explainable artificial intelligence (XAI) techniques in patients with intracranial hemorrhage: A systematic review," *Wiley Interdisciplinary Reviews: Data Mining and Knowledge Discovery,* vol. 15, no. 3, p. e70031, 2025.

[7] Z. Wu *et al.*, "AgentGraph: Trace-to-Graph Platform for Interactive Analysis and Robustness Testing in Agentic AI Systems," in *Proceedings of the AAAI Conference on Artificial Intelligence*, 2026, vol. 40, no. 48, pp. 41721-41723.

[8] D. Wang, W. Wei, L. Li, X. Wang, and J. Liang, "Rethinking exploration–exploitation trade-off in reinforcement learning via cognitive consistency," *Neural Networks,* vol. 187, p. 107342, 2025.

[9] S. Huang, C. Sun, R.-Q. Wang, and D. Pompili, "Toward adaptive and coordinated transportation systems: A multi-personality multi-agent meta-reinforcement learning framework," *IEEE Transactions on Intelligent Transportation Systems,* 2025.

[10] Y. Zhai *et al.*, "Agentevolver: Towards efficient self-evolving agent system," *arXiv preprint arXiv:2511.10395,* 2025.

[11] M. Xu, Y. Mei, F. Zhang, and M. Zhang, "Genetic programming and reinforcement learning on learning heuristics for dynamic scheduling: A preliminary comparison," *IEEE Computational Intelligence Magazine,* vol. 19, no. 2, pp. 18-33, 2024.

[12] K. Hirasawa, M. Okubo, H. Katagiri, J. Hu, and J. Murata, "Comparison between Genetic Network Programming (GNP) and Genetic Programming (GP)," in *Proceedings of the 2001 Congress on Evolutionary Computation*, 2001 2001, vol. 2, pp. 1276-1282 vol. 2, doi: 10.1109/CEC.2001.934337.

[13] M. Roshanzamir, R. Alizadehsani, S. V. Moravvej, J. H. Joloudari, H. Alinejad-Rokny, and J. M. Gorriz, "Enhancing Interpretability in Machine Learning: A Focus on Genetic Network Programming, Its Variants, and Applications," in *Artificial Intelligence for Neuroscience and Emotional Systems*, Cham, J. M. Ferrández Vicente, M. Val Calvo, and H. Adeli, Eds., 2024// 2024: Springer Nature Switzerland, pp. 98-107.

[14] X. Li, B. Li, S. Mabu, and K. Hirasawa, "A novel estimation of distribution algorithm using graph-based chromosome representation and reinforcement learning," in *IEEE Congress of Evolutionary Computation*, 5-8 June 2011 2011, pp. 37-44, doi: 10.1109/CEC.2011.5949595.

[15] X. Li, S. Mabu, and K. Hirasawa, "Use of infeasible individuals in probabilistic model building genetic network programming," in *13th annual conference on Genetic and evolutionary computation*, Dublin, Ireland, 2011, 2001659: ACM, pp. 601-608, doi: 10.1145/2001576.2001659.

[16] X. Li, S. Mabu, and K. Hirasawa, "An extended probabilistic model building genetic network programming using both of good and bad individuals," *IEEJ Transactions on Electrical and Electronic Engineering,* vol. 8, no. 4, pp. 339-347, 2013, doi: 10.1002/tee.21864.
[17] K. Shimada, "Genetic network programming with acquisition mechanisms of association rules," *Journal of Advanced Computational Intelligence and Intelligent Informatics,* vol. 10, no. 1, pp. 102-111, 2006. [Online]. Available: http://ci.nii.ac.jp/naid/20000779642/en/.
[18] S. Mabu, T. Higuchi, and T. Kuremoto, "SemiSupervised Learning for Class Association Rule Mining Using Genetic Network Programming," *IEEJ Transactions on Electrical and Electronic Engineering,* vol. 15, no. 5, pp. 733-740, 2020, doi: https://doi.org/10.1002/tee.23109.
[19] Y. Xu, Y. Sun, Z. Ma, H. Zhao, Y. Wang, and N. Lu, "Attribute Selection Based Genetic Network Programming for Intrusion Detection System," *Journal of Advanced Computational Intelligence and Intelligent Informatics,* vol. 26, no. 5, pp. 671-683, 2022, doi: 10.20965/jaciii.2022.p0671.
[20] S. Mabu, C. Chen, N. Lu, K. Shimada, and K. Hirasawa, "An Intrusion-Detection Model Based on Fuzzy Class-Association-Rule Mining Using Genetic Network Programming," *Transactions on Systems, Man, and Cybernetics,* vol. 41, no. 1, pp. 130-139, 2011, doi: 10.1109/tsmcc.2010.2050685.
[21] N. Lu, S. Mabu, T. Wang, and K. Hirasawa, "Integrated fuzzy GNP rule mining with distance-based classification for intrusion detection system," in *IEEE International Conference on Systems, Man, and Cybernetics*, 14-17 Oct. 2012 2012, pp. 1569-1574, doi: 10.1109/ICSMC.2012.6377960.
[22] S. Nahavandi, "Industry 5.0—A human-centric solution," *Sustainability,* vol. 11, no. 16, p. 4371, 2019.
[23] M. Ghobakhloo, H. A. Mahdiraji, M. Iranmanesh, and V. Jafari-Sadeghi, "From Industry 4.0 digital manufacturing to Industry 5.0 digital society: a roadmap toward human-centric, sustainable, and resilient production," *Information Systems Frontiers,* pp. 1-33, 2024.
[24] A. Akundi, D. Euresti, S. Luna, W. Ankobiah, A. Lopes, and I. Edinbarough, "State of Industry 5.0—Analysis and identification of current research trends," *Applied System Innovation,* vol. 5, no. 1, p. 27, 2022.
[25] X. Xu, Y. Lu, B. Vogel-Heuser, and L. Wang, "Industry 4.0 and Industry 5.0—Inception, conception and perception," *Journal of manufacturing systems,* vol. 61, pp. 530-535, 2021.
[26] J. Yang, Y. Liu, and P. L. Morgan, "Human-machine interaction towards Industry 5.0: Human-centric smart manufacturing," *Digital Engineering,* p. 100013, 2024.
[27] E. Gonzales, K. Shimada, S. Mabu, K. Hirasawa, and J. Hu, "Genetic network programming with parallel processing for association rule mining in large and dense databases," in *Proceedings of the 9th annual conference on Genetic and evolutionary computation*, 2007, pp. 1512-1512.
[28] H. Madokoro, S. Nix, and K. Sato, "Automatic Calibration of Piezoelectric Bed-Leaving Sensor Signals Using Genetic Network Programming Algorithms," *Algorithms,* vol. 14, no. 4, p. 117, 2021. [Online]. Available: https://www.mdpi.com/1999-4893/14/4/117.
[29] R. Ramezanian, A. Peymanfar, and S. B. Ebrahimi, "An integrated framework of genetic network programming and multi-layer perceptron neural network for prediction of daily stock return: An application in Tehran stock exchange market," *Applied Soft Computing,* vol. 82, p. 105551, 2019/09/01/ 2019, doi: https://doi.org/10.1016/j.asoc.2019.105551.
[30] F. Köhnke and C. Borgelt, "Variable-Size Genetic Network Programming for Portfolio Optimization with Trading Rules," in *International Conference on the Applications of Evolutionary Computation (Part of EvoStar)*, 2025: Springer, pp. 287-304.
[31] R. S. Sutton and A. G. Barto, *Reinforcement learning: An introduction* (no. 1). A Bradford Book, 2018.
[32] Q. Meng, S. Mabu, and K. Hirasawa, "Genetic Network Programming with Sarsa Learning Based Nonuniform Mutation," in *IEEE International Conference on Systems, Man and Cybernetics*, 10-13 Oct. 2010 2010, pp. 1273-1278, doi: 10.1109/ICSMC.2010.5642421.
[33] P. Sung Gil, S. Mabu, and K. Hirasawa, "Robust Genetic Network Programming using SARSA Learning for autonomous robots," in *ICCAS-SICE*, 18-21 Aug. 2009 2009, pp. 523-527.
[34] S. Mabu, H. Hatakeyama, K. Hirasawa, and H. Jinglu, "Genetic Network Programming with Reinforcement Learning Using Sarsa Algorithm," in *IEEE International Conference on Evolutionary Computation*, 0-0 0 2006, pp. 463-469, doi: 10.1109/CEC.2006.1688346.
[35] Y. Chen, S. Mabu, K. Hirasawa, and J. Hu, "Trading rules on stock markets using genetic network programming with sarsa learning," presented at the Proceedings of the 9th annual conference on Genetic and evolutionary computation, London, England, 2007.
[36] Y. Chen, S. Mabu, and K. Hirasawa, "Genetic Network Programming with Reinforcement Learning and Its Application to Creating Stock Trading Rules," in *Machine Learning*: InTech, 2009.

[37] X. Li and K. Hirasawa, "A Learning Classifier System Based on Genetic Network Programming," in *IEEE International Conference on Systems, Man, and Cybernetics*, 13-16 Oct. 2013 2013, pp. 1323-1328, doi: 10.1109/SMC.2013.229.

[38] X. Li, M. Yang, and S. Wu, "Niching genetic network programming with rule accumulation for decision making: An evolutionary rule-based approach," *Expert Systems with Applications,* vol. 114, pp. 374-387, 2018, doi: https://doi.org/10.1016/j.eswa.2018.07.041.

[39] X. Li, S. Mabu, H. Zhou, K. Shimada, and K. Hirasawa, "Genetic Network Programming with Estimation of Distribution Algorithms for class association rule mining in traffic prediction," in *IEEE Congress on Evolutionary Computation*, 18-23 July 2010 2010, pp. 1-8, doi: 10.1109/CEC.2010.5586456.

[40] X. Li, S. Mabu, and K. Hirasawa, "A Novel Graph-Based Estimation of the Distribution Algorithm and its Extension Using Reinforcement Learning," *IEEE Transactions on Evolutionary Computation,* vol. 18, no. 1, pp. 98-113, 2014, doi: 10.1109/TEVC.2013.2238240.

[41] X. Li, B. Li, S. Mabu, and K. Hirasawa, "A continuous estimation of distribution algorithm by evolving graph structures using reinforcement learning," in *2012 IEEE Congress on Evolutionary Computation*, 10-15 June 2012 2012, pp. 1-8, doi: 10.1109/CEC.2012.6256481.

[42] X. Li, W. He, and K. Hirasawa, "Adaptive Genetic Network Programming," in *IEEE Congress on Evolutionary Computation*, 6-11 July 2014 2014, pp. 1808-1815, doi: 10.1109/CEC.2014.6900290.

[43] M. Roshanzamir, M. Palhang, and A. Mirzaei, "Graph structure optimization of Genetic Network Programming with ant colony mechanism in deterministic and stochastic environments," *Swarm and Evolutionary Computation,* vol. 51, p. 100581, 2019, doi: https://doi.org/10.1016/j.swevo.2019.100581.

[44] M. Roshanzamir, M. Palhang, and A. Mirzaei, "Efficiency improvement of genetic network programming by tasks decomposition in different types of environments," *Genetic Programming and Evolvable Machines,* vol. 22, no. 2, pp. 229-266, 2021/06/01 2021, doi: 10.1007/s10710-021-09402-y.

[45] M. Roshanzamir and M. Roshanzamir, "Chapter Four - Situation-based genetic network programming to solve agent control problems," in *Advances in Computers*, vol. 135, A. Biswas, A. P. Tonda, R. Patgiri, and K. K. Mishra Eds.: Elsevier, 2024, pp. 77-97.

[46] A. Kohan, M. Roshanzamir, and R. Alizadehsani, "Enhancing Multiagent Genetic Network Programming Performance Using Search Space Reduction," in *International Conference on the Dynamics of Information Systems*, 2025: Springer, pp. 94-105.

[47] Y. Zhang, X. Li, Y. Yang, S. Mabu, Y. Jin, and K. Hirasawa, "Functionally distributed systems using parallel Genetic Network Programming," in *Proceedings of SICE Annual Conference 2010*, 18-21 Aug. 2010 2010, pp. 2626-2630.

[48] Y. Liu, G. Zhang, K. Wang, S. Li, and S. Pan, "Graph-augmented large language model agents: Current progress and future prospects," *arXiv preprint arXiv:2507.21407,* 2025.

[49] Y. Bei *et al.*, "Graphs meet ai agents: Taxonomy, progress, and future opportunities," *arXiv preprint arXiv:2506.18019,* 2025.

[50] Y. Wu *et al.*, "AgentKit: structured LLM reasoning with dynamic graphs," *arXiv preprint arXiv:2404.11483,* 2024.

[51] C. Huan *et al.*, "Scaling graph chain-of-thought reasoning: A multi-agent framework with efficient llm serving," *arXiv preprint arXiv:2511.01633,* 2025.

[52] J. Gao *et al.*, "Graph counselor: Adaptive graph exploration via multi-agent synergy to enhance llm reasoning," in *Proceedings of the 63rd Annual Meeting of the Association for Computational Linguistics (Volume 1: Long Papers)*, 2025, pp. 24650-24668.

[53] Y. Yang, J. Tang, L. Xia, X. Zou, Y. Liang, and C. Huang, "Graphagent: Agentic graph language assistant," in *Proceedings of the 2025 Conference on Empirical Methods in Natural Language Processing*, 2025, pp. 26360-26379.

[54] X. Wu *et al.*, "Can graph learning improve planning in llm-based agents?," *Advances in Neural Information Processing Systems,* vol. 37, pp. 5338-5383, 2024.

[55] Y. Hu, R. Lei, X. Huang, Z. Wei, and Y. Liu, "Scalable and accurate graph reasoning with llm-based multi-agents," *arXiv preprint arXiv:2410.05130,* 2024.

[56] Z. Zhang *et al.*, "AgentRouter: A Knowledge-Graph-Guided LLM Router for Collaborative Multi-Agent Question Answering," *arXiv preprint arXiv:2510.05445,* 2025.

[57] G. Zhang *et al.*, "Evoflow: Evolving diverse agentic workflows on the fly," *arXiv preprint arXiv:2502.07373,* 2025.

[58] S. Prasai, M. Du, Y. Zhang, and F. Yang, "KnowThyself: An Agentic Assistant for LLM Interpretability," in *Proceedings of the AAAI Conference on Artificial Intelligence*, 2026, vol. 40, no. 48, pp. 41661-41663.

[59] H. Katagiri, K. Hirasawa, H. Jinglu, and J. Murata, "Comparing some graph crossover in genetic network programming," in *Proceedings of the 41st SICE Annual Conference. SICE 2002.*, 5-7 Aug. 2002 2002, vol. 2, pp. 1263-1268 vol.2, doi: 10.1109/SICE.2002.1195369.

[60] X. Li, W. He, and K. Hirasawa, "Genetic Network Programming with Simplified Genetic Operators," in *Neural Information Processing: 20th International Conference, ICONIP 2013, Daegu, Korea, November 3-7, 2013. Proceedings, Part II*, M. Lee, A. Hirose, Z.-G. Hou, and R. M. Kil Eds. Berlin, Heidelberg: Springer Berlin Heidelberg, 2013, pp. 51-58.

[61] X. Li, H. Yang, and M. Yang, "Revisiting genetic network programming (gnp): towards the simplified genetic operators," *IEEE Access,* vol. 6, pp. 43274-43289, 2018.

[62] E. G. Liquin and A. Gopnik, "Children are more exploratory and learn more than adults in an approach-avoid task," *Cognition,* vol. 218, p. 104940, 2022.

[63] S. Kim and S. M. Carlson, "Understanding explore-exploit dynamics in child development: current insights and future directions," *Frontiers in Developmental Psychology,* vol. 2, p. 1467880, 2024.

[64] W. E. Frankenhuis and A. Gopnik, "Early adversity and the development of explore–exploit tradeoffs," *Trends in Cognitive Sciences,* vol. 27, no. 7, pp. 616-630, 2023.

[65] M. Pollack and M. Ringuette, "Introducing the Tileworld: experimentally evaluating agent architectures," *environment,* pp. 183-189, 1990.

[66] D. D. Wackerly, W. Mendenhall, and R. L. Scheaffer, *Mathematical statistics with applications*. Thomson Brooks/Cole Belmont, CA, 2008.

[67] D. Cheng *et al.*, "Reasoning with exploration: An entropy perspective," in *Proceedings of the AAAI Conference on Artificial Intelligence*, 2026, vol. 40, no. 36, pp. 30377-30385.